\documentclass[12pt,journal,onecolumn,draftclsnofoot]{IEEEtran}

\usepackage[utf8]{inputenc} 

\usepackage[T1]{fontenc}    
\usepackage{hyperref}       
\usepackage{url}            
\usepackage{booktabs}       
\usepackage{amsfonts}       
\usepackage{nicefrac}       
\usepackage{microtype}      
\usepackage{xcolor}         

\usepackage{enumerate}

\usepackage{graphicx}
\usepackage{subfigure}
\usepackage{amsmath, amsthm, amssymb}
\usepackage{algorithm}

\usepackage{footnote}
\makesavenoteenv{algorithm}

\usepackage{bbm}
\usepackage{amsfonts}
\usepackage{amssymb}
\usepackage{color,soul}
\usepackage{float}
\usepackage{soul}
\usepackage{changes}
\usepackage[noend]{algpseudocode}
\usepackage{wrapfig}
\usepackage{scalerel}
\usepackage[T1]{fontenc}

\newtheorem{theorem}{Theorem}
\newtheorem{lemma}{Lemma}
\usepackage{epstopdf}
\usepackage{epsfig}
\usepackage{tikz}

\usepackage{mathtools}
\DeclarePairedDelimiter\ceil{\lceil}{\rceil}
\DeclarePairedDelimiter\floor{\lfloor}{\rfloor}

\usepackage{soul}

\theoremstyle{plain}
\newtheorem{thm}{Theorem}
\newtheorem{cor}[thm]{Corollary}

\newtheorem{prop}[thm]{Proposition}

\newtheorem{rem}{Remark}

\newtheorem*{defin*}{Definition}

\newtheorem{assum}{Assumption}
\newcommand{\twopartdef}[4]
{
	\left\{
	\begin{array}{lll}
		#1 & \mbox{ } #2 \\
		#3 & \mbox{ } #4 
	\end{array}
	\right.
}

\newcommand{\qname}{QuBan}

\newcommand{\lin}[1]{\textcolor{red}{[Lin: #1]}}

\begin{document}

%

%

\title{Solving Multi-Arm Bandit Using a Few Bits of Communication} 

\author{\IEEEauthorblockN{Osama A. Hanna$^\dagger$, Lin F. Yang$^\dagger$ and Christina Fragouli$^\dagger$\\ 
$^\dagger$University of California, Los Angeles\\
Email:\{ohanna, linyang, christina.fragouli\}@ucla.edu}
}
\maketitle

\begin{abstract}
The multi-armed bandit (MAB) problem is an active learning framework that aims to select the best among a set of actions by sequentially observing rewards. Recently, it has become popular for a number of applications over wireless networks, where communication constraints can form a bottleneck. Existing works usually fail to address this issue and can become infeasible in certain applications. In this paper we address the communication problem by optimizing the communication of rewards collected by distributed agents. 
By providing nearly matching upper and lower bounds, we tightly characterize the \emph{number of bits} needed per reward for the learner to accurately learn without suffering additional regret.
In particular, we 
establish a generic reward quantization algorithm, $\qname$, that can be applied on top of any (no-regret) MAB algorithm to form a new communication-efficient counterpart, that requires only a few (as low as $3$) bits to be sent per iteration while preserving the \emph{same} regret bound. Our lower bound is established via constructing hard instances from a subgaussian distribution. Our theory is further corroborated by numerically experiments.

\end{abstract}
\section{Introduction}
Multi-armed bandit (MAB) is an active learning framework that finds applications in diverse domains, including recommendation systems, clinical trials, adaptive routing, and so on \cite{bouneffouf2019survey}.
In a MAB problem, a learner interacts with an environment by pulling an arm from a set of arms, each of which, if played, gives a scalar reward, sampled from an unknown but fixed distribution. 
The goal of the learner is to find the arm with the highest mean reward using a minimum number of pulls.
The performance of a learner is measured in terms of regret, that captures the expected difference between the observed rewards and rewards drawn from the best arm.
Work on MAB algorithms and their applications spans several decades, cultivating a rich literature that considers a variety of models and algorithmic approaches \cite{lattimore2020bandit}. {MAB algorithms include explore-then-commit \cite{robbins1952some, anscombe1963sequential}, $\epsilon$-greedy \cite{auer2002finite}, Thomson sampling \cite{thompson1933likelihood}, and the upper confidence bound (UCB) \cite{lai1987adaptive, auer2002finite}, to name a few. Under some assumptions on the reward distribution, the explore-then-commit and $\epsilon$-greedy algorithms achieve a  regret bound $\propto O(\sqrt{n})$, where $n$ is number of steps the learner plays, for the worst-case but known minimum reward gap\footnote{The reward gap is defined to be the difference between the reward means of the best and second best arm.}, while Thomson sampling and UCB achieve a regret bound $\propto O(\sqrt{n\log(n)})$ without knowledge of the minimum means gap}\footnote{{Variants of the UCB \cite{audibert2009minimax, degenne2016anytime} can achieve regret  $\propto O(\sqrt{n})$, but can be worse than UCB in some regimes \cite{lattimore2020bandit}.}}. However, all these works assume that the rewards can be communicated to the learner at full precision which can be costly in communication constrained setups.  In this paper we ask: \emph{is it possible to perform efficient and effective bandit learning with only a few bits communicated per reward?}

Understanding how many bits of communication are really needed, is not only interesting from a theoretical viewpoint, but can also enable the MAB framework to support learning applications in settings that were challenging before.  Consider, for instance,  swarms of tiny robots (such as RoboBees and RoboFlies \cite{wood2013flight}), wearable (inside and outside the body) sensors, backscatterer and RFID networks, IoT and embedded systems; generally whenever low complexity sensors cooperate, the communication cost can become a performance bottleneck for any learning framework.  MAB systems in areas such as  mobile healthcare, social decision-making and spectrum allocation  have already been implemented in a distributed manner, using limited bandwidth wireless links and simple sensors with low computational power \cite{anandkumar2011distributed, buccapatnam2013multi, buccapatnam2014stochastic, mary2015bandits, song2018recommender, ding2019interactive}; reducing the number of bits directly translates to reduced power consumption and wireless interference for these systems.  

\begin{figure}
\centering
 \includegraphics[width=0.7\linewidth]{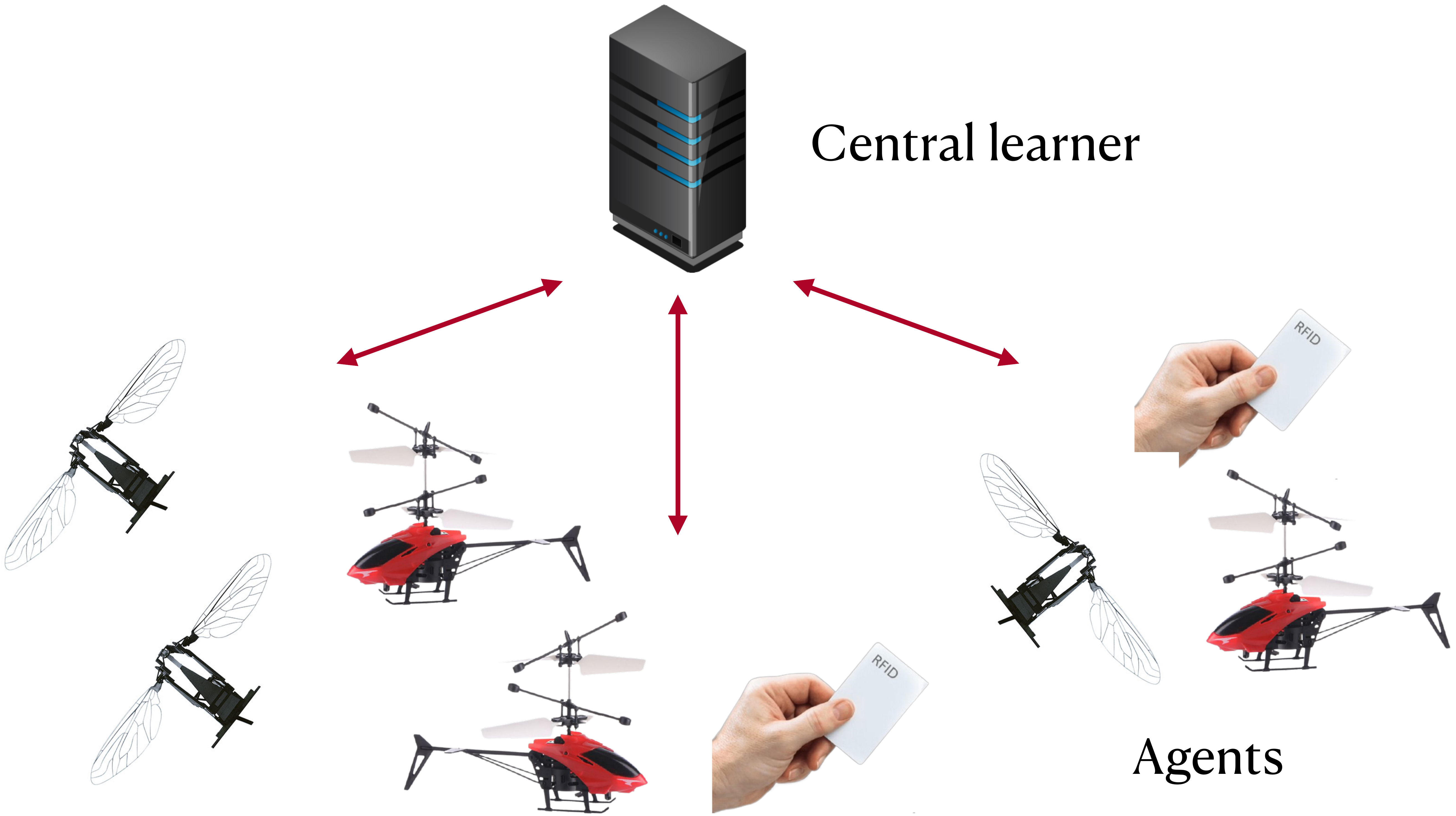}
  \caption{A central learner collects rewards from a set of agents. The agents can join and leave at any time and hence can be different and unaware of the historical rewards, i.e., memoryless.}\label{fig:model}
\end{figure}

In this paper we consider a common setup 
 illustrated in Fig.~\ref{fig:model}, where a central learner can directly communicate with a set of agents. We assume that the agents may change from time to time (e.g., are mobile), but that each agent can pull whichever arm  the learner requests it to, 
observe the reward,
and  communicate the reward to the learner. 
For example,
the learner 
could be a "traffic policeman" for small drones that searches best current policies; or a base-station that helps low-capability sensors achieve spectrum sharing.
For many existing systems, the learner may have already implemented a MAB algorithm to handle the learning task. 
Hence our goal is to design a communication protocol such that the rewards are communicated with only a few bits and yet the performance of the original MAB algorithm does not degrade.


Our main contribution is a set of upper and lower bounds on the required number of bits to achieve the same unquantized regret up to a small constant factor. In particular, our lower bound states that we need to send at least $2.2$ bits per reward to {maintain properties of the reward that enable to} achieve a regret within  a factor of $1.5$ from the unquantized regret.  This lower bound is established through proving necessary properties of the optimal quantizer, then constructing a hard instance from a gaussian distribution.
Our upper bounds state that, on average, $3.4$ bits are sufficient to maintain {the required properties and achieve} a regret within a factor of $1.5$ from the unquantized regret  (and that we rarely need to use more than $4$ bits).

 The upper bounds are proved using a novel quantization scheme, that we term $\qname$, tailored to compressing MAB rewards. 
$\qname$ only cares to maintain what matters to the MAB algorithm operation, namely the ability to decide which is the best arm. 
At a high level, $\qname$ maps  rewards to quantization levels chosen to be dense around an estimate of the arm's mean values and  sparse otherwise. $\qname$ employs a stochastic correction term  that enables to convey an unbiased estimate of the rewards with a small variance. $\qname$ introduces a simple novel rounding trick to guarantee that the quantization error is conditionally independent on the history given the current pulled arm index. This maintains the Markov property which is crucial in the analysis of bandit algorithms and enables reusing the same analysis methods for unquantized rewards to bound the regret after quantization. Finally, $\qname$ encodes the reward values that occur more frequently with shorter representations, in order to reduce the number of bits communicated.
We provide empirical studies for a number of MAB algorithms, e.g.,
UCB and $\epsilon$-greedy. 
Numerical results corroborate that $\qname$,  applied on top of MAB algorithms, using a few bits (as small as $3$) can achieve the same regret as unquantized communication.\\
\textbf{Related Work.} 
To the best of our knowledge, the proposed model is novel and no scheme from the literature can be used to solve the problem of maintaining a regret bound that matches the unquantized regret bound, up to a small constant factor, while using a few bits of communication. In the following, we distinguish our work from a representative sample of existing literature.\\
\noindent{\em MAB algorithms.}
There is a long line of research in the literature about MAB algorithms. 
For instance, explore-then-commit \cite{robbins1952some, anscombe1963sequential}, $\epsilon$-greedy \cite{auer2002finite}, Thomson sampling \cite{thompson1933likelihood}, the upper confidence bound \cite{lai1987adaptive, auer2002finite} and its variant for contextual bandits \cite{dani2008stochastic, li2010contextual}. {Under the assumption that the reward distributions are $1$-subgaussian, these algorithms provide a worst case regret that is almost $O(\sqrt{n})$. The explore-then-commit regret is upper bounded by $C\sqrt{n}$ for bandits with $2$ arms and known minimum means gap, while the regret of $\epsilon$-greedy is upper bounded by $C'\sqrt{kn}$ for $k$-armed bandits with knowledge of the minimum means gap, where $C,C'$ are constants that do not depend on $k,n$ \cite{lattimore2020bandit}. Thomson sampling and UCB achieve a regret with upper bound $C\sqrt{kn\log(n)}$ for $k$-armed bandits, where $C$ is a constant that does not depend on $k,n$ \cite{russo2014learning, agrawal2012analysis, agrawal2013further, katehakis1995sequential, agrawal1995sample,auer2002finite}. Contextual Thomson sampling and LinUCB achieve a regret bounded by $Cd\sqrt{n}\log(n)$, where $d$ is the dimension of an unknown system parameter, and $C$ is a universal constant that does not depend on $n,d$ \cite{agrawal2013thompson, abeille2017linear, lattimore2020bandit, dani2008stochastic}.}
These algorithms assume access to a full precision reward at each iteration. Our goal is not replacing existing MAB algorithms to deal with quantized rewards; instead, we are interested in a general quantization framework that can be applied on top of any existing (or future) MAB algorithm.

\noindent{\em Compression for ML and distributed optimization.}
There is a number of research results targeting reducing the communication cost of learning systems using compression.
For instance, compression is applied on gradient updates \cite{seide20141, alistarh2017qsgd, mayekar2020limits}. 
Recent work has also looked at compression for classification tasks \cite{hanna2020distributed}. 
However, compression schemes tailored to active learning, such as MAB problems, have not been explored.
Our quantization scheme can be understood as a reward compression scheme that reduces the communication complexity for MABs. The main difference between the quantization for MABs and for distributed learning is that the later targets reducing the dependency of the number of bits and performance on the dimensionality of bounded training data, which can be in the order of tens of millions. In contrast, the rewards of MABs are scalars. The main challenge of our setting is to deal with a reward distribution that is either unbounded or the upper bound on the reward is much larger than the noise variance, which is typical in many MAB applications. This can be done by exploiting the fact that the rewards are more likely to be picked from the arm that appear to be best. Such a property is not applicable in the general distributed optimization setup and comes with new challenges as will discussed later.

\noindent{\em Sample complexity.} Compression is related to sample complexity \cite{even2002pac, mannor2004sample}: indeed, sending a small number of samples, reduces the overall communication load.
However, the question we ask is different (and complementary): sample complexity asks how many (full precision) samples from each distribution do we need to draw; we are asking, how many bits of each  sample do we really need to transmit, when we only care to decide the best arm and not to reconstruct the samples.

\noindent{\em Distributed multi-agent MAB.} Researchers have explored  the distributed multi-agent MAB problem with a single  \cite{anantharam1987asymptotically} or multiple \cite{shahrampour2017multi} decision makers; in these settings, distributed agents pick arms under some constraints (all agents pick the same arm \cite{shahrampour2017multi}, at most one agent can pick the same arm at a time otherwise no reward is given \cite{anandkumar2011distributed} and other constraints \cite{landgren2019distributed}). The agents cooperate to aggregate their observed rewards so as to jointly make a more informed decision on the best arm. Most of the works do not take into account communication constraints, and rather focus on cooperation/coordination schemes. Our setup is different: we have a single learner (central server) and simple agents who do not learn (do not keep memory) but simply observe and transmit rewards, one at a time. Our scheme can be potentially applied to these settings to reduce communication cost.

Independently and in parallel to ours, the work in \cite{vial2020one} 
also considered MAB learning with reduced number of bits, restricted in their case to UCB policies.
Their main result shows that for rewards supported on $[0,1]$, one bit of communication is sufficient; our work recovers this result using a much simpler approach as a special case of  Section~\ref{sec:example}. Additionally, our work applies on top of any MAB algorithm, and for unbounded rewards.

\noindent\textbf{Paper Organization.} 
Section~\ref{sec:model} presents our model and notation; Section~\ref{sec:example} looks at a special case; Section~\ref{sec:alg} describes $\qname$; Section~\ref{sec:per} presents our main theorems and  Section~\ref{sec:eval} provides numerical evaluation.

\section{Model and Notation}\label{sec:model}
\textbf{MAB Framework.} We consider a multi-armed bandit (MAB) problem over a horizon of size $n$ \cite{robbins1952some}. At each iteration  $t=1,...,n$, a learner chooses an arm (action) $A_t$ from a set of arms $\mathcal{A}_t$ and receives a random reward $r_t$ distributed according to an unknown reward distribution with mean $\mu_{A_t}$. {The reward distributions are assumed to be $\sigma^2$-subgaussian   \cite{boucheron2013concentration}}. The arm selected at time $t$ depends on the previously selected arms and observed rewards $A_1,r_1,...,A_{t-1},r_{t-1}$. The learner is interested in minimizing the expected regret $R_n=\mathbb{E}[R'_n]$, where $R'_n$ is the regret defined as
\begin{equation}\label{eq:regret}
    R'_n=\Sigma_{t=1}^n(\mu^*_t-r_t),
\end{equation}
where $\mu^*_t=\max_{A\in \mathcal{A}_t}\mu_{A}$. 
The expected regret captures the difference between the expected total reward collected by the learner over $n$ iterations and the reward if we selected the arm with the maximum mean (optimal arm). 

\noindent{\em Notation. } {When the set of arms $\mathcal{A}_t$ is finite and does not depend on $t$: we denote the number of arms by $k=|\mathcal{A}_t|$, the best arm mean by $\mu^*$, and the gap between the best arm and the arm-$i$ mean by $\Delta_i:=\mu^*-\mu_i$.} {If $X,Y$ are random variables, we refer to the expectation of $X$, variance of $X$, conditional expectation of $X$ given $Y$, and conditional variance of $X$ given $Y$ as $\mathbb{E}[X]$, $\sigma^2(X)$, $\mathbb{E}[X|Y]$, and $\sigma^2(X|Y)$ respectively.}

Popular MAB algorithms for the case where the set of actions is fixed over time, $\mathcal{A}=\mathcal{A}_t$, and $\mathcal{A}$ is finite include explore-then-commit \cite{robbins1952some, anscombe1963sequential}, $\epsilon$-greedy \cite{auer2002finite}, Thomson sampling \cite{thompson1933likelihood}, and UCB \cite{lai1987adaptive, auer2002finite}. In addition to this case we also consider an important class of bandit problems, contextual bandits \cite{langford2007epoch, agrawal2013thompson}. In this case, before picking an action, the learner observes a side information, the context. Specifically we consider the widely used stochastic linear bandits model \cite{abe1999associative}, where the contexts are modeled by changing the action set $\mathcal{A}_t$ across time. In this model, at iteration $t$, the learner chooses an action $A_t$ from a given set $\mathcal{A}_t\subseteq \mathbb{R}^d$ and gets a reward
\begin{equation}
    r_t=\langle \theta_*,A_t \rangle+\eta_t,
\end{equation}
where $\theta_*\in \mathbb{R}^d$ is an unknown parameter, and $\eta_t$ is a noise.
Conditioned on $\mathcal{A}_1,A_1,r_1,...,\mathcal{A}_t,A_t,r_t$, the noise $\eta_{t+1}$ is assumed to be zero mean {and $\sigma^2$-subgaussian}. Popular algorithms for this case include LinUCB \cite{dani2008stochastic}, explore-then-commit strategy \cite{rusmevichientong2010linearly}, and contextual Thomson sampling \cite{agrawal2013thompson}.

\textbf{System Setup.} We are interested in a distributed setting, where a learner
asks at each time a potentially different agent to play the arm $A_t$; the agent observes the reward $r_t$ and conveys it to the learner over a communication constrained channel, as depicted in Fig.~\ref{fig:model}. In our setup, each agent needs to immediately communicate the observed reward (with no memory), using a quantization scheme to reduce the communication cost. As learning progresses, the learner is allowed to refine the quantization scheme by broadcasting parameters to the agents they may need. We do not count these broadcast (downlink) transmissions in the communication cost since the learner has no restrictions in its power.  
We stress again that the agents cannot store information of the reward history since they may \emph{join and leave} the system at any time.
We thus opt to use a setting where the agents have no memory.
This setting allows to
support 
applications with simple agents (e.g. RFID applications and embedded systems).

\textbf{Quantization.} 
{A quantizer consists of an encoder \mbox{$\mathcal{E}:\mathbb{R}\to \mathcal{S}$}  that maps  $\mathbb{R}$ to a countable set $\mathcal{S}$, and a decoder $D:\mathcal{S}\to \mathbb{R}$}. At each time $t$, the agent that observes the reward $r_t$ transmits {a finite length binary sequence representing $\mathcal{E}(r_t)$} {to the learner which in turn decodes it using the decoder $D$ to obtain the quantized reward $\hat{r}_t=D(\mathcal{E}(r_t))$.}
The range of a decoder is {referred to as} the set of quantization levels; the end-to-end operation of a quantizer maps the reward to a quantization level.
We next describe a specific quantization module that we will use.

\textbf{Stochastic Quantization (SQ).}
A stochastic quantizer that uses quantization levels in a set $\mathcal{L}$, which is a form of dithering \cite{gray1993dithered, alistarh2017qsgd}, consists of a randomized encoder $\mathcal{E}_\mathcal{L}$ and decoder $D_\mathcal{L}$ modules that can be described as following. The encoder $\mathcal{E}_\mathcal{L}$, {that uses the set of quantization levels $\mathcal{L}=\{\ell_i\}_{i=1}^{2^B}$}, takes as input a value $x$ in $[\ell_1,\ell_{2^B}]$; it maps $x$ to a level index described by $B$ bits. The decoder, {that uses the set of quantization levels $\mathcal{L}=\{\ell_i\}_{i=1}^{2^B}$}, takes as input an index in $\{1,...,2^B\}$, and outputs the corresponding level value. Precisely,
\begin{align} \label{eq:enc-dec}
 &i(x)=\max \{j| \ell_j\leq x \; \text{and}\; j<2^B\},\nonumber \\
    &\mathcal{E}_\mathcal{L}(x)=\twopartdef
{i(x)}      {\text{ with probability } \frac{\ell_{i(x)+1}-x}{\ell_{i(x)+1}-\ell_{i(x)}}}
{i(x)+1}     {\text{ with probability } \frac{x-\ell_{i(x)}}{\ell_{i(x)+1}-\ell_{i(x)}}},\nonumber \\
& D_\mathcal{L}(j) = \ell_j, j\in \{1,...,2^B\}.
\end{align}
That is, if $x$ is such that $\ell_i\leq x < \ell_{i+1}$, then the index $i$ is transmitted with probability $\frac{\ell_{i+1}-x}{\ell_{i+1}-\ell_i}$ (and $x$ is decoded to be $\ell_i$) while the index $i+1$ is transmitted with probability $\frac{x-\ell_{i}}{\ell_{i+1}-\ell_i}$ (and $x$ is decoded to be $\ell_{i+1}$).

The analysis of  bandit algorithms leverages the fact that conditioned on $A_t$, the communicated reward $r_t$ is an unbiased estimate of the mean $\mu_{A_t}$. {It is not difficult to see that} SQ preserves this property, namely conditioned on $A_t$, it conveys to the learner an unbiased estimate of $\mu_{A_t}$.

\textbf{Performance Metric $B_n, \Bar{B}(n)$.} Among the schemes that achieve a regret matching the unquantized regret, {up to a fixed small constant factor}, {our performance metrics are the instantaneous and average number of communication bits per reward $B_n$, and $\Bar{B}(n)$ respectively.}
Let $B_t$ be the number of bits used to transmit $\hat{r}_t$, and define the average number of bits after $n$ iterations of the algorithm as $\Bar{B}(n)=\frac{\sum_{t=1}^nB_t}{n}$. 
Our goal is to design quantization schemes that achieve expected regret matching the expected regret of unquantized communication  (up to a small constant factor) while using a  small number of bits $B_n$, and $\Bar{B}(n)$.

\section{A Case Where $1$ Bit is Sufficient} \label{sec:example}
In this section we consider the special case where the rewards are supported on $[0,1]$. For simplicity, we also assume that all reward distributions have the same range and the same variance $\sigma$ (but different means).

We explore schemes that use exactly one bit per reward; note that one bit is a trivial lower bound on the number of bits communicated, since each agent needs to respond to the learner for each observed reward.  
In particular, we use  $1$-bit Stochastic Quantization (SQ),
as in \eqref{eq:enc-dec}.  
Assume that $r_t\in [0,1]$ and the variance $\sigma^2(r_t|A_t) \approx 1/4$. The stochastic $1$ bit quantizer  takes $r_t$ as input and interprets it as probability:  outputs $1$ with probability $r_t$ and $0$ with probability $1-r_t$. Let $\hat{r}_t$ be the (binary) quantized reward, we then have that
\begin{equation}
    \mathbb{E}[\hat{r}_t|A_t]=\mathbb{E}[\mathbb{E}[\hat{r}_t|r_t,A_t]|A_t]=\mathbb{E}[r_t|A_t]=\mu_{A_t}.
\end{equation}
Since $\hat{r}_t\in[0,1]$, its variance  is upper bounded by $\frac{1}{4}$. Recall that for bandit algorithms the expected regret scales linearly with the variance. For example, the UCB algorithm (c.f., \cite{lattimore2020bandit}) with unquantized rewards, achieves $R_n\leq C \sqrt{nk\log(n)}$ for a constant $C$ that does not depend on $k$, $n$. It is not difficult to see that similar to \cite{auer2002finite}, UCB with  $1$-bit SQ achieves a regret bound $R_n\leq C \sqrt{nk\log(n)}$. 
Simulation results verify that, for $r_t\in [0,1]$, 1-bit SQ performs very close to  unquantized rewards  (proofs and simulation results are in App.~\ref{app:mot} and Section~\ref{sec:eval}).

To motivate our general quantization scheme, we consider a case where $1$-bit SQ results in a potentially large performance loss.
Assume that the variance, $\sigma$, is much smaller than the range of $r_t$:  $r_t\in [-\lambda,\lambda]$ and $\sigma=1$, where $\lambda\gg 1$ is a parameter known to the learner.
The $1$-bit SQ maps $r_t$ to either $\lambda$ or $-\lambda$; it is not difficult to see that  we still have $\mathbb{E}[\hat{r}_t|A_t]=\mu_{A_t}$, but $R_n\leq C\lambda \sqrt{kn\log(n)}$, where $C$ is a constant that does not depend on $n,k$ \cite{auer2002finite}\footnote{We note that this bound cannot be improved using techniques in \cite{auer2002finite}, since it is possible that $\sigma^2(\hat{r}_t|A_t)=\lambda^2$ (e.g., if $r_t=0$ almost surely).}. Thus the expected regret bound grows linearly with $\lambda$, which can be arbitrarily large.
In contrast, without quantization UCB still achieves $C'\sqrt{kn\log(n)}$, where $C'$ is another constant of the same order of $C$.
Simulation results verify  that the convergence to the unquantized case can be slow.

We take away the following observations:\\
$\bullet$  If the range $\lambda$ is of the same order as the variance $\sigma$, 1-bit SQ is sufficient to preserve the regret bound up to a small constant factor.\\
$\bullet$ If the range $\lambda$ is much larger than $\sigma$, 1-bit SQ leads to a regret {penalty} proportional to {$\frac{\lambda}{\sigma}$}; thus we may want to only perform stochastic quantization within intervals of size similar to $\sigma$.

In this section we assumed that the rewards $r_t$ are bounded almost surely. This is not true in general; we would like an algorithm that uses a small average number of bits even when the reward distributions are unbounded; $\qname$, described next, achieves this. 

\section{$\qname$: A General  Quantizer for  Bandit Rewards} \label{sec:alg}
In this section, we propose $\qname$, an adaptive quantization scheme that can be applied on top of any MAB algorithm. Our scheme maintains attractive properties (such that the Markov property, unbiasedness, and bounded variance) for the quantized rewards that enable to retain the same regret bound as unquantized communication for the vast majority of MAB algorithms, while using a few bits for communication (simulation results show convergence to $\sim 3$ bits per iteration for $n$ that is sufficiently large, see Section~\ref{sec:eval}). We use ideas that include: (i) centering the quantization scheme around a value that is believed to be close to the picked arm mean in the majority of iterations; (iii) maintaining a quantization error that is conditionally independent on previously observed rewards given the arm selection, which is achieved by choosing the quantization center to be an integer value (illustrated in more detail in the proof of Theorem 2); (iv) assigning shorter codes to the values near the quantization center and otherwise longer codes to maintain a finite expected number of bits even if the reward distribution has infinite support; and (iv) using stochastic quantization to convey an unbiased estimate of the reward.

$\qname$ builds on the following observations. Recall that at time $t$ the learner selects an action $A_t$ and needs to convey the observed reward $r_t$.
As we expect $r_t$ to be close to the mean $\mu_{A_t}$, we would like to use quantization levels that are dense around $\mu_{A_t}$  and sparse in other areas.
\begin{figure}[t!]
  \centering
     \includegraphics[width=0.9\linewidth]{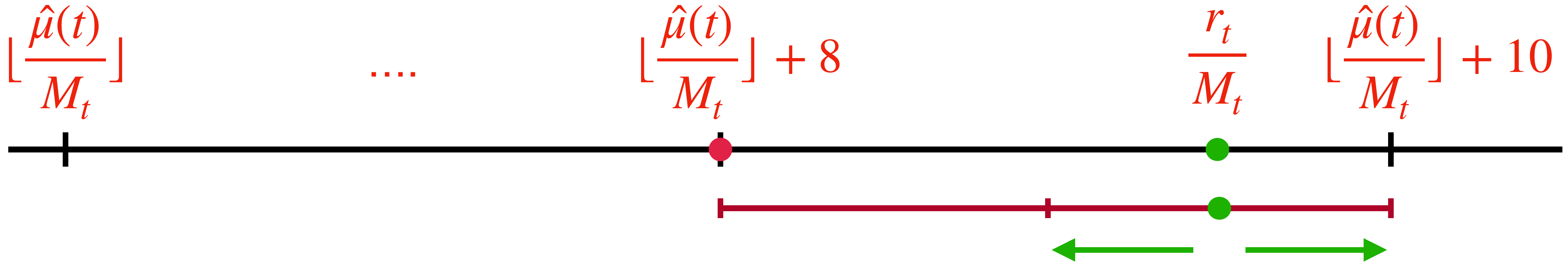}
  \caption{Illustration of $\qname$. In the shown example, $r_t$ is mapped to a value of the red dot (conveyed with the index $I_t=4$), and stochastically to one of the two nearest quantization levels depicted on the red line.}\label{fig:quant}
\end{figure}
Since $\mu_{A_t}$ is unknown, we estimate it using some function of the observed rewards that we term  $\hat{\mu}(t)$; 
we can think of  $\hat{\mu}(t)$ as specifying a ``point" on the real line  around which we  want to provide denser quantization.

\subsection{Choices for $\hat{\mu}(t)$}
In this work, we analyze the following three choices for $\hat{\mu}(t)$, the first two applying to MAB with a finite {fixed set} of arms, while the third to linear bandits.\\
\noindent $\bullet$ \textbf{Average arm point (Avg-arm-pt):} $\hat{\mu}(t)=\hat{\mu}_{A_t}(t-1)$. We use $\hat{\mu}_{A_t}(t-1)$, the average of the samples picked from arm $A_t$ up to time $t-1$, as an estimate of $\mu_{A_t}$.\\
\noindent $\bullet$ \textbf{Average point (Avg-pt)}: $\hat{\mu}(t)=\frac{1}{t-1}\sum_{j=1}^{t-1}\hat{r}_j$ (the average over all observed rewards). Here we can think of $\frac{1}{t-1}\sum_{j=1}^{t-1}\hat{r}_j$ as an estimate of the mean of the best arm. Indeed, the average reward of a well behaved algorithm will converge to the best mean reward.

These two choices of $\hat{\mu}(t)$ give us flexibility to fit different regimes of MAB systems.
In particular, we expect the {avg-arm-pt} to be a better choice for a small number of arms $k$ and MAB algorithms that achieve good estimates of $\mu_{A_t}$ (explore all arms sufficiently so that $\hat{\mu}_{A_t}(t-1)$ approaches $\mu_{A_t}$). However, as our analysis also shows (see Section~\ref{sec:per}), if $k$ is large, acquiring good estimates for all arms may be costly and not what good algorithms necessarily pursue; instead, the {avg-pt}
has a simpler implementation, as it only requires to keep track of a single number, and still enables to distinguish well in the neighborhood of the best arm, which is essentially what we mostly want.\\
\noindent{$\bullet$} \textbf{Contextual bandit choice}: $\hat{\mu}(t)=\langle \theta_t,A_t \rangle$. 
Consider the widely used stochastic linear bandits model  in Section~\ref{sec:model}. We observe that {linear bandit algorithms}, such as contextual Thomson sampling and LinUCB, choose a parameter $\theta_t$ believed to be close to the unknown parameter $\theta_*$, and pick an action based on $\theta_t$. For example, LinUCB \cite{dani2008stochastic} chooses a confidence set $\mathcal{C}_t$ with center $\theta_t$ believed to contain $\theta_*$ and picks an action $A_t=\arg\max_{a\in \mathcal{A}_t}\max_{\theta\in \mathcal{C}_t}\langle \theta, a\rangle$.
Accordingly, we propose to use $\hat{\mu}(t)=\langle \theta_t,A_t \rangle$.
We note that our intuition for the {avg-pt} choice does not work for contextual bandits as it relies on that $\max_{a\in \mathcal{A}_t}\langle \theta_*,a\rangle$ is the same for all $t$, which might not hold in general. Likewise, the {avg-arm-pt} choice will not work as the set of actions $\mathcal{A}_t$ can be infinite or change with time. 

We underline that the  estimator $\hat{\mu}(t)$ is only maintained at the learner's side and is broadcasted to the agents. As discussed before, this downlink communication is not counted as communication cost.

\subsection{$\qname$ Components}
At iteration $t$, $\qname$ centers its quantization around the value $\hat{\mu}(t)$. It then quantizes the normalized reward $\Bar{r}_t={r_t}/{M_t}-\floor{{\hat{\mu}(t)}/{M_t}}$ to one of the two values $\floor{\Bar{r}_t}, \ceil{\Bar{r}_t}$, where $M_t=\epsilon \sigma X_t$\footnote{The case where $\sigma$ is unkown is discussed in App.~\ref{app:disc}.}, $\epsilon$ is a parameter to control the regret vs number of bits trade-off as will be illustrated later in this section, and $\{X_t\}_{i=1}^n$ are independent samples from a $\frac{1}{4}$-subgaussian distribution satisfying $|X_t|\geq 1$ almost surely, e.g., we can use $X_t=1$ almost surely\footnote{For our proofs we set $X_t=1$ for simplicity; more sophisticated choices can further improve the upper bounds as discussed in App.~\ref{app:disc}.}. This introduces an error in estimating $\Bar{r}_t$ that is bounded by $1$, which results in error of at most $M_t$ in estimating $r_t=M_t (\Bar{r}_t+\floor{{\hat{\mu}(t)}/{M_t}})$. This quantization is done in a randomized way to convey an unbiased estimate of $r_t$. 

\begin{algorithm}[tb]
  \caption{Learner operation with input MAB algorithm $\Lambda$}
  \label{alg:quant}  
  \begin{algorithmic}[1]
  \State Initialize: $\hat{\mu}(1)=0$
  \For{$t=1,...,n$}
    \State Choose an action $A_t$ based on the bandit\\ \quad\quad algorithm {${\Lambda}$} and ask the next agent to play it
    \State Send $M_t$\footnote{If $X_t$ is chosesn to be $1$, then sending $M_t$ is not required.}, $\hat{\mu}(t)$ to an agent
    \State Receive the encoded reward $(b_t,I_t,\mathcal{E}_{\mathcal{L}_t}(e_t))$ (see\\ \quad\quad Algorithm~\ref{alg:quant_reward})
    \State {\bf Decode $\hat{r}_{t}$:} \If{length($b_t$)$\leq 4$} 
    \State $\hat{r}_t$ can be decoded using a lookup table
    \Else
    \State Decode the sign, $s_t$, of $r_t$ from $b_t$
    \State Set $\ell_t$ to be the $I_t$-th element in the set\\ \quad\quad\quad $\{0,2^0,...\}$
    \State Set $\mathcal{L}_t=\{\ell_t, \ell_t+1,...,\max\{2\ell_t,\ell_t+1\}\}$
    \State Let $e_{t}^{(q)}=D_{\mathcal{L}_{t}}(\mathcal{E}_{\mathcal{L}_{t}}(e_{t}))$
    \State $\hat{r}_{t}= (s_t(e_{t}^{(q)}+\ell_t+3.5)+0.5+\lfloor {\hat{\mu}(t)}/{M_t}\rfloor)M_t$
    \EndIf
    \State {\bf Calculate $\hat{\mu}(t+1)$} (using one of the discussed\\ \quad\quad choices)
    \State {{\bf Update the parameters required by ${\Lambda}$}}
  \EndFor
  \end{algorithmic}
\end{algorithm}
\begin{algorithm}[t]
  \caption{Distributed Agent Operation}
  \label{alg:quant_reward}  
  \begin{algorithmic}[1]
    \item \textbf{Inputs:} $r_t$, $\hat{\mu}(t)$ and $M_t$
    \State Set $L=\{\floor{\Bar{r}_t},\ceil{\Bar{r}}\}$, $\hat{\Bar{r}}_t=D_L(\mathcal{E}_L(\Bar{r}_t))$
    \State Set $b_t$ with three bits to distinguish between the $8$ cases: $\hat{\Bar{r}}_t<-2, \hat{\Bar{r}}_t>3, \hat{\Bar{r}}_t=i, i\in \{-1,0,1,2\}$.
    \If{$|\hat{\Bar{r}}_t|>|a|$ and $\hat{\Bar{r}}_ta>0$, $a\in \{-2,3\}$}
    \State Augment $b_t$ with an extra one bit to indicate if $|\hat{\Bar{r}}_t|=|a|+1$ or $|\hat{\Bar{r}}_t|>|a|+1$.
    \If{$|\hat{\Bar{r}}_t|>|a|+1$}
        \State Let $L'=\{0,2^0,...\}$
        \State Set $\ell_t=\max\{j\in L| j\leq |\Bar{r}_t|-|a|\}$
        \State Encode $\ell_t$ by $I_t-1$ zeros followed by a one\\ \quad\quad\quad (unary coding), where $I_t$ is the index of $\ell_t$\\ \quad\quad\quad in the set $L'$.
        \State Let $e_t=|\Bar{r}_t|-|a|-\ell_t$
        \State Set $\mathcal{L}_t=\{\ell_t, \ell_t+1,...,\max\{2\ell_t,\ell_t+1\}\}$
        \State Encode $e_t$ using SQ to get $\mathcal{E}_{\mathcal{L}_t}(e_t)$
    \EndIf
  \EndIf
  \State Transmit $(b_t,I_t,\mathcal{E}_{\mathcal{L}_t}(e_t))$
  \end{algorithmic}
\end{algorithm}
{\bf Rounding of ${\hat{\mu}(t)}/{M_t}$:} the reason for choosing the quantization to be centered around $\floor{{\hat{\mu}(t)}/{M_t}}$ instead of ${\hat{\mu}(t)}/{M_t}$ is to guarantee that the distance between $r_t$ and the two closest quantization levels is independent of $\hat{\mu}(t)$\footnote{{As will be shown in App.~\ref{app:prop}, centering the quantization around any integer value implies that the two closest quantization levels to $\frac{r_t}{M_t}$ are $\floor{\frac{r_t}{M_t}}, \ceil{\frac{r_t}{M_t}}$}.} (which is dependent on $\hat{r}_1,...,\hat{r}_{t-1}$). {As we discuss in the following section, this preserves the Markov property (given $A_t$, the quantized reward $\hat{r}_t$ is conditionally independent on the history $A_1,\hat{r}_1,...,A_{t-1},\hat{r}_{t-1}$), a property that is exploited in the analysis of bandit algorithms to guarantee that  $|{\sum_{t=1}^n\hat{r}_t-\mu_{A_t}}/{n}|$ approaches zero in some probabilistic sense as $n$ increases.}

The precise learner and agent operations used for $\qname$ are presented in pseudo-code in Algorithms~\ref{alg:quant} and \ref{alg:quant_reward} (see Fig.~\ref{fig:quant} for an example), respectively, and discussed in detail in App.~\ref{app:disc}, \ref{app:main_thm}.
The learner at each iteration broadcasts $\hat{\mu}(t)$  and asks one of the agents {available at time $t$} to play an action $A_t$.
Initially, since we have no knowledge about $\mu_i$, the learner assumes that $\hat{\mu}(0)=0$.  The agent that plays the action  uses the
observed $r_t$ together with $\hat{\mu}(t)$ it has received to transmit three values we term  $(b_t,I_t,e_t)$, to the learner, as described in Algorithm 2 using $O(\log(|\Bar{r}_t|))$ bits.  
\section{Main Results}\label{sec:per}
Our main results provide an upper and lower bound on the number of bits required to achieve the same unquantized regret up to a small constant factor. In particular, we show that $2.2$ bits per reward are required to achieve a sub-linear regret and a quantization error, $\hat{r}_t-r_t$, that is $(\frac{\sigma}{2})^2$ subgaussian\footnote{The subgaussian condition is required for the standard analysis techniques of many algorithms, which is why we want to satisfy this property for the quantized rewards.}. These conditions imply a regret within a factor of $1.5$ from the unquantized regret. We also show that, on the average, $3.4$ bits are sufficient to maintain a $(\frac{\sigma}{2})^2$-subgaussian quantization error, and achieve a regret within a factor of $1.5$ from the unquantized regret. Before stating the results, we state our assumptions.

\begin{assum}
\label{assum:mab}
We assume that all the codes are prefix free \cite{cover1999elements}\footnote{A similar analysis can be carried out for non-singular codes \cite{cover1999elements}.} and that we are given:\\
(i) a MAB instance with $\sigma^2$-subgaussian\footnote{This is a standard assumption used for simplicity but is not required for our main results.} rewards where the Markov property holds: conditioned on the action at time $t$, the current reward is conditionally independent on the history (past actions and rewards).\\
(ii) a MAB algorithm $\Lambda$ such that for any instance with $\sigma^2$-subgaussian rewards, and time horizon $n$, the algorithm's expected regret (with unquantized rewards) is upper-bounded by $R_n^U$.
\end{assum}

The following proposition gives an upper bound on the regret after quantization showing that for $\epsilon=1$, the regret is within a factor of $1.5$ from the regret of the unquantized case. The proof is provided in App.~\ref{app:prop}.

\begin{prop}\label{prop:1}
 Suppose Assumption~\ref{assum:mab} holds. Then, when we apply $\qname$, the following hold:\\
    \noindent$1.$ Conditioned on $A_t$, the quantized reward $\hat{r}_t$ is $((1+\frac{\epsilon}{2})\sigma)^2$-subgaussian, conditionally independent on the history $A_1,\hat{r}_1,...,A_{t-1},\hat{r}_{t-1}$ (Markov property), and satisfies
        $\mathbb{E}[\hat{r}_t|A_t]=\mu_{A_t},\  |\hat{r}_t-r_t|\leq M_t \text{ almost surely}$ ($t=1,\ldots, n$).\\
    \noindent$2.$ The expected regret $R_n$  is bounded as  
    $R_n\leq (1+\frac{\epsilon}{2})R_n^U$, where $\epsilon$ is a parameter to control the regret vs number of bits trade-off.
\end{prop}

In the following we provide upper bound on the expected average number of bits. We also provides a high-probability upper bound on the instantaneous number of bits. For simplicity we only consider the case where $\epsilon=1$ and discuss the other case in App.~\ref{app:main_thm}. The proof is given in App.~\ref{app:main_thm}. 

\begin{theorem}\label{main_thm}
Suppose Assumption~\ref{assum:mab} holds. Let $\epsilon=1$. There is a universal constant $C$ such that:\\
    \noindent$1.$ For $\qname$ with $\hat{\mu}(t)=\hat{\mu}_{A_t}(t-1)$ (avg-arm-pt),  the average number of bits communicated satisfies that $\mathbb{E}[\Bar{B}(n)]\leq 3.4+({C}/{n}){\sum_{i=1}^k \log(1+{|\mu_i|}/{\sigma})}+{C}/{\sqrt{n}}$.\\
    \noindent$2.$ For $\qname$ with $\hat{\mu}(t)=\frac{1}{t-1}\sum_{j=1}^{t-1}\hat{r}_j$ (avg-pt),  the average number of bits communicated satisfies
    $\mathbb{E}[\Bar{B}(n)]\leq 3.4+\frac{C}{n}\left({1+ \log(1+\frac{|\mu^*|}{\sigma})+\frac{R_n}{\sigma}+\sum_{t=1}^{n-1}\frac{R_t}{(\sigma t)}}\right)+{C}/{\sqrt{n}}$.\\
     \noindent$3.$ For  $\qname$ with $\hat{\mu}(t)=\langle \theta_t,A_t \rangle$ (stochastic linear bandit), the average number of bits communicated satisfies that $\mathbb{E}[\Bar{B}](n)\leq 3.4+C{\mathbb{E}[\sum_{t=1}^{n}|\langle \theta_t-\theta_*, A_t \rangle|]}/{(\sigma n)}$.\\
\end{theorem}
In App.~\ref{app:main_thm} we also provide almost surely bounds on the asymptotic average number of bits, namely, $\lim_{n\to \infty}({1}/{n})\sum_{t=1}^nB_t{\leq} 3.4$ almost surely.

In the following we provide a high probability bound on the number of bits that $\qname$ uses in each iteration. We analyze the performance for avg-arm-pt only; the other choices for $\hat{\mu}(t)$ can be handled similarly.
\begin{theorem}
For a MAB instance with $\sigma^2$-subgaussian rewards, $\qname$ with $\epsilon =1, \hat{\mu}(t)=\hat{\mu}_{A_t}(t-1)$ (avg-arm-pt), satisfies that for $t$ with $T_t(A_t)>0$, where $T_t(i)$ is the number of pulls for arm $i$ prior to iteration $t$, with probability at least $1-\frac{1}{n}$ it holds that $\forall t\leq n$:
\begin{equation}\label{eq:inst}
    B_t\leq 4+\ceil{\log(4\log(n))}+\ceil{\log\log(4\log(n))}.
\end{equation}
\end{theorem}
The proof is provided in App.~\ref{app:high_prob}.

\begin{rem}
Using the previous lemma we can modify $\qname$ to have that \eqref{eq:inst} is satisfied almost surely, by sending a random $1$ bit when \eqref{eq:inst} is not satisfied. This will only add at most $n\sum_{i=1}^k \Delta_i$ regret with probability at most $\frac{1}{n}$. Hence, the expected regret is increased by at most a factor of $2$.
\end{rem}

\subsection{Lower Bound}
In this subsection we provide a lower bound showing that an average number of $2.2$ bits per iteration are required to maintain a sublinear regret and a $(\frac{\sigma}{2})^2$-subgaussian quantization error, $\hat{r}_t-r_t$. We also show that the instantaneous number of bits cannot be almost surely bounded by a constant.
\begin{theorem}\label{thm:lower}
For any memoryless algorithm that only uses quantized rewards, prefix-free encoding and satisfies that for any MAB instance with subgaussian rewards:\\
\noindent$i.$ $R_n$ is sublinear in $n$,\\
\noindent$ii.$ Conditioned on $r_t$, $\hat{r}_t-r_t$ is $(\frac{\sigma}{2})^2$-subgaussian ($t=1,\ldots, n$),\\ we have that there exist $\sigma^2$-subgaussian reward distributions for which:\\
\noindent$1.$ $(\forall b\in \mathbb{N})$ $(\exists t,\delta>0)$ such that $\mathbb{P}[B_t>b]>\delta$.\\
\noindent$2.$ $(\forall t>0)$ $(\exists n>t)$ such that $\mathbb{E}[\Bar{B}(n)]\geq 2.2$ bits.
\end{theorem}
Our lower bound is established by proving necessary properties for the set of quantization schemes that satisfy $i,ii$ which include that $\mathbb{E}[\hat{r}_t|r_t]=r_t$ and that the distance between the quantization levels cannot be too large. We then show that $2.2$ bits are needed to satisfy the proved properties for a Gaussian distribution. The proof is given in App.~\ref{app:lower}.

\subsection{Application to UCB, $\epsilon$-greedy, and LinUCB}
We here leverage Theorem~\ref{main_thm}  to derive bounds for three widely used  MAB algorithms. We highlight that although the regret bounds hide constant factors, these constants are within $1.5$ of the unquantized constants according to Theorem~\ref{main_thm}.
The proofs are in App.~\ref{app:cor} for Corollaries~\ref{cor1} and~\ref{cor2} and in App.~\ref{app:contex_bandits} for Lemma~\ref{lem1}.
\begin{cor} \label{cor1}
Assume we use $\qname$ with avg-pt on top of \textbf{UCB} \cite{auer2002finite} with $\sigma^2$-subgaussian reward distributions and worst case gap  $\Delta_i$. Then there is a constant $C$ that does not depend on $n$ and $k$ such that $R_n\leq C\sigma \sqrt{nk\log(n)},\ \mathbb{E}[\Bar{B}(n)]\leq 3.4+C\sqrt{{k\log(n)}/{n}}$.
\end{cor}
\begin{cor}\label{cor2}
Assume we use $\qname$ with avg-pt on top of \textbf{$\epsilon$-greedy} \cite{auer2002finite} with $\sigma^2$-subgaussian reward distributions and constant gaps  $\Delta_i$ $\forall i$. Let $\epsilon_t=\min\{1,{Ck}/{(t\Delta_{\text{min}}^2})\}$, where $\Delta_{\text{min}}=\min_{i}\{\Delta_i|\Delta_i>0\}$ and $C>0$ is a sufficiently large universal constant. Then there exists a constant $C'$ that does not depend on $n$ and $k$ such that $R_n\leq C'\sigma k\log(1+{n}/{k}),\ \mathbb{E}[\Bar{B}(n)]\leq 3.4+C'({k\log^2(n)}/{n}+{1}/{\sqrt{n}}).$
\end{cor}

To simplify the expressions, we include the dependency on $\mu^*$ and $\Delta_i$ in the constant $C$ for Corollary~\ref{cor1} and respectively $C'$ for Corollary~\ref{cor2}.
\begin{lemma}\label{lem1} Assume we use $\qname$ on top of \textbf{LinUCB} \cite{dani2008stochastic}.
Under the assumptions in App.~\ref{app:contex_bandits}, there is a constant $C$ that does not depend on $n$ and $d$ such that $R_n\leq Cd\sqrt{n}\log(n),\ \mathbb{E}[\Bar{B}(n)]\leq 3.4+C\frac{d\log(n)}{\sqrt{n}}$.
\end{lemma}

\section{Numerical Evaluation} \label{sec:eval}
\begin{figure*}[t!]
  \centering
\subfigure[Setup 1 {(larger $\Delta_i$ values).}]{\includegraphics[width=0.32\linewidth]{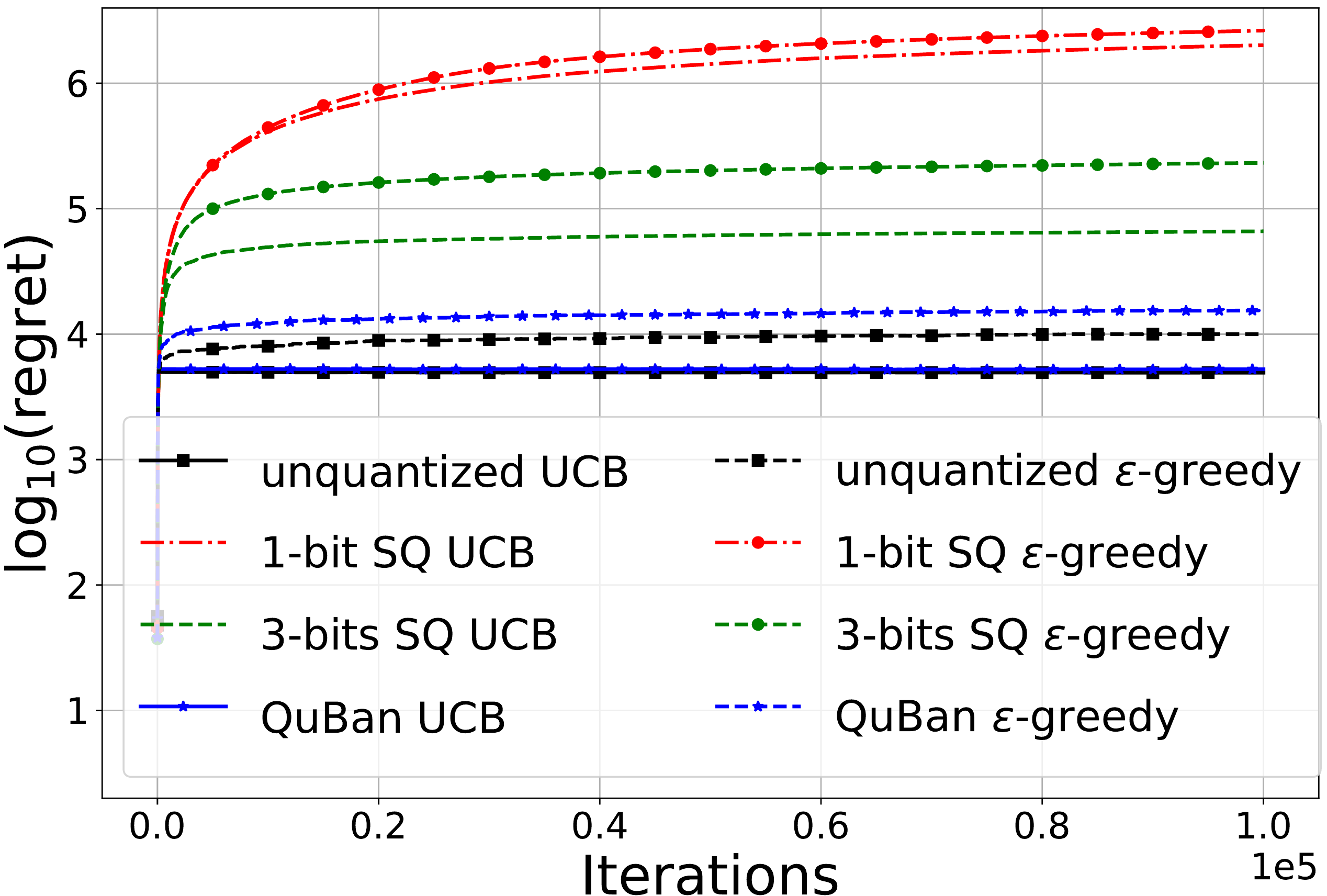}\label{fig:regrets_all}}
 \subfigure[Setup 2 {(smaller $\Delta_i$ values).}]{\includegraphics[width=0.32\linewidth]{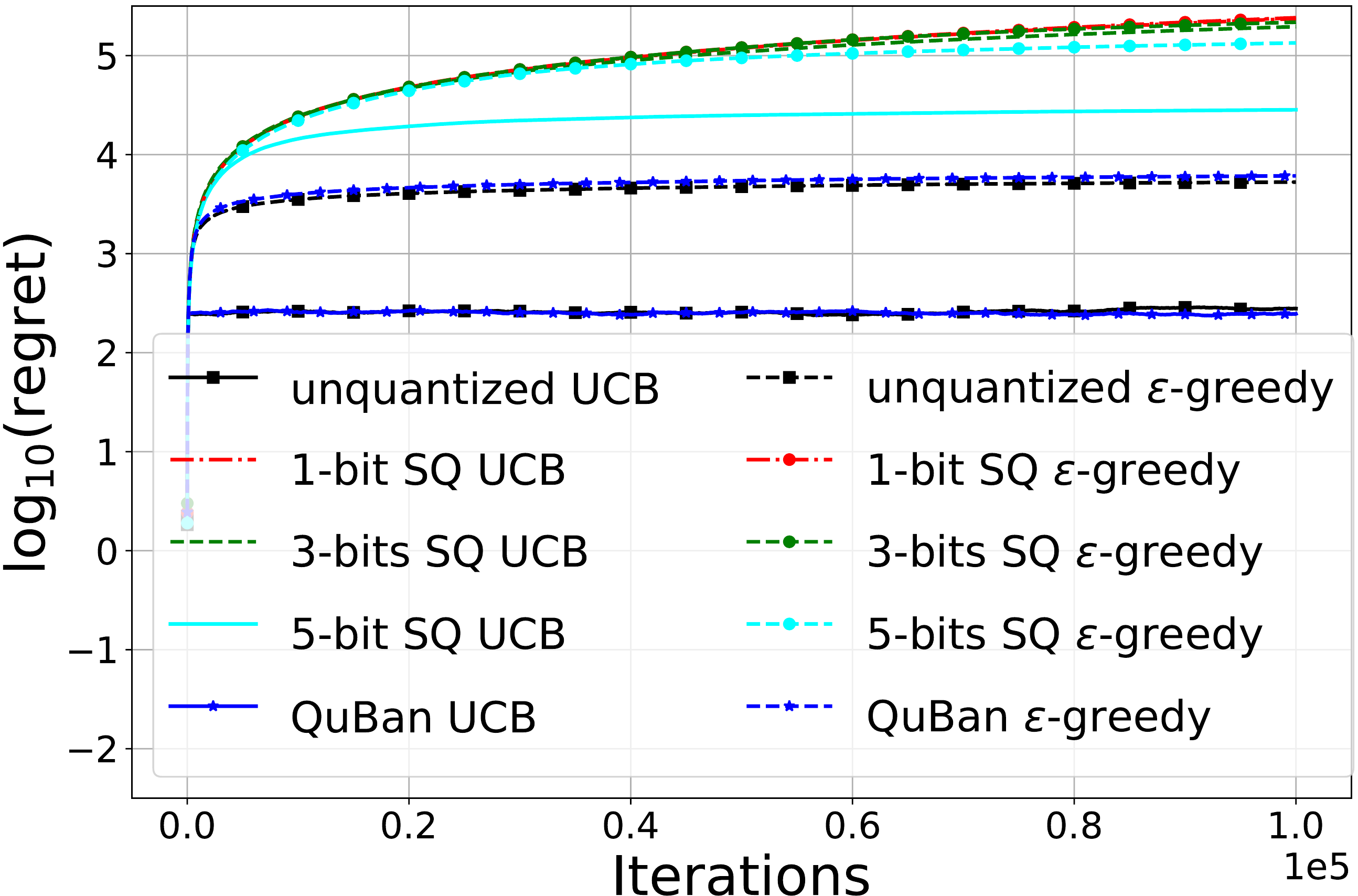}\label{fig:regret_UCB}}
  \subfigure[Setup 3 {(linear bandits).}]{\includegraphics[width=0.32\linewidth]{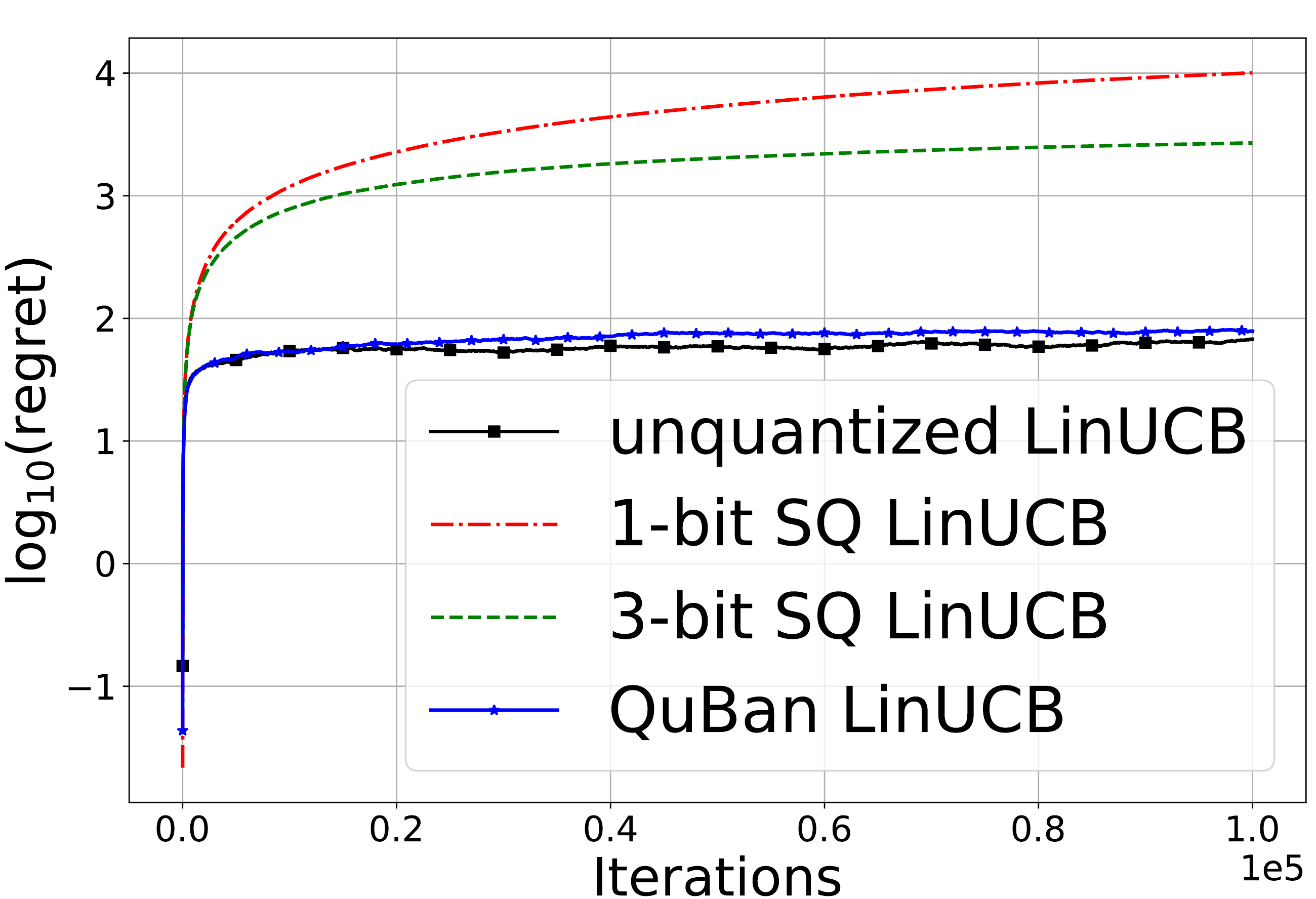}\label{fig:regret_LinUCB}}
  \vspace{-3mm}
  \caption{Regret versus number of iterations. }\label{fig:iter}
  \vspace{-5mm}
\end{figure*}

\begin{figure*}[t!]
  \centering
\subfigure[Setup 1 {(larger $\Delta_i$ values).}]{\includegraphics[width=0.32\linewidth]{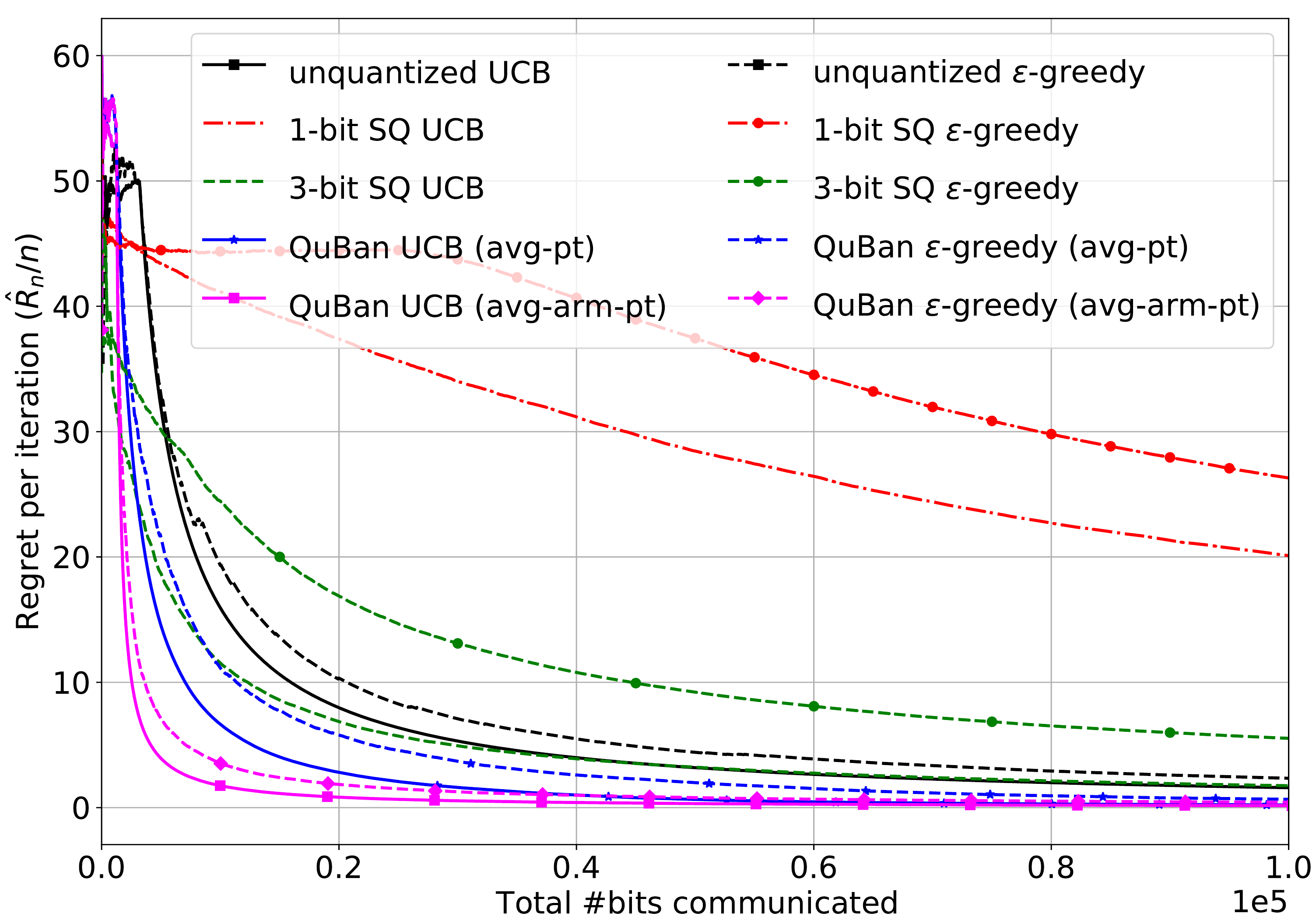}\label{fig:regrets_bits}}
  \subfigure[Setup 2 {(smaller $\Delta_i$ values).}]{\includegraphics[width=0.32\linewidth]{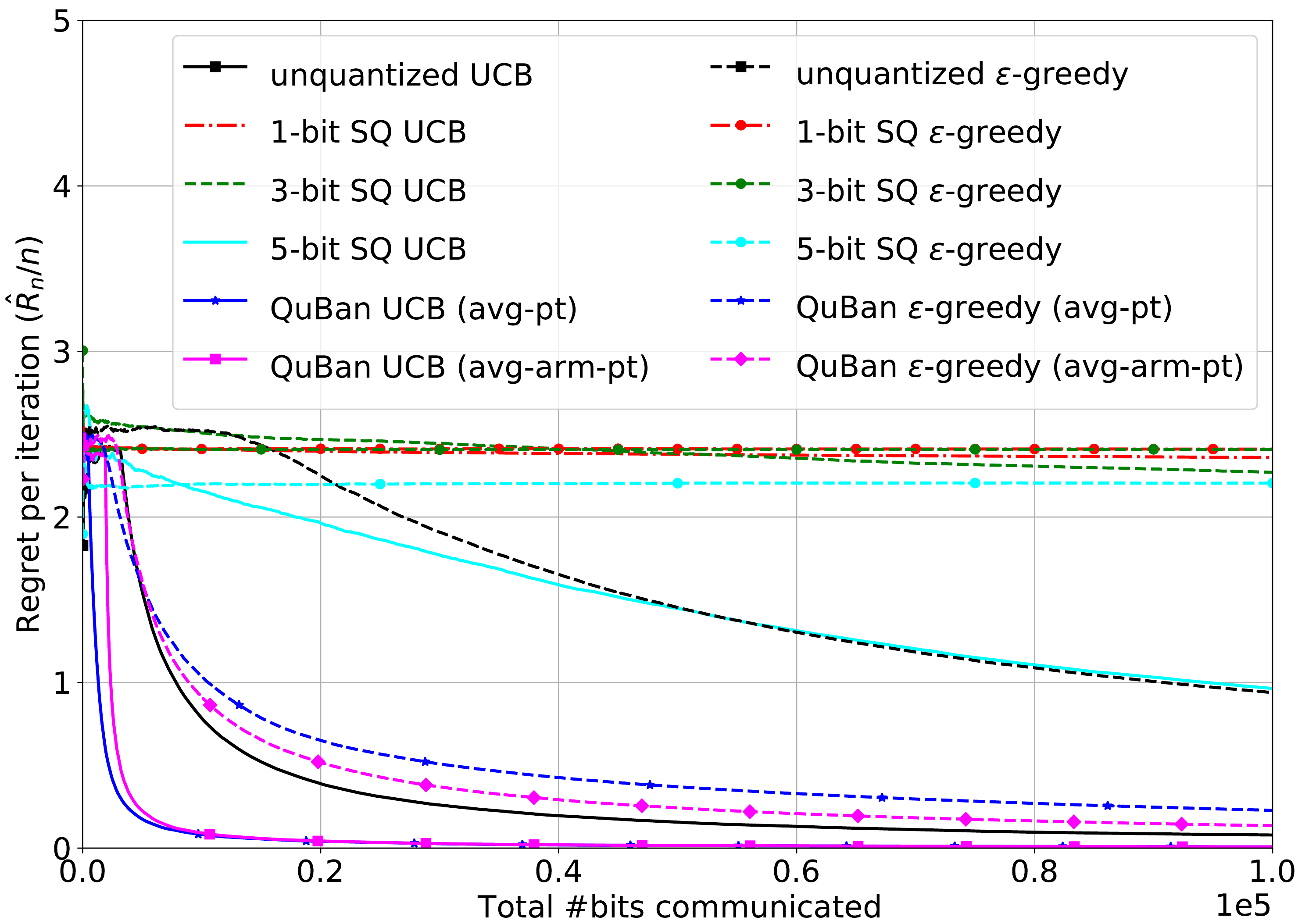}\label{fig:bits_UCB}}
  \subfigure[Setup 3 {(linear bandits).}]{\includegraphics[width=0.32\linewidth]{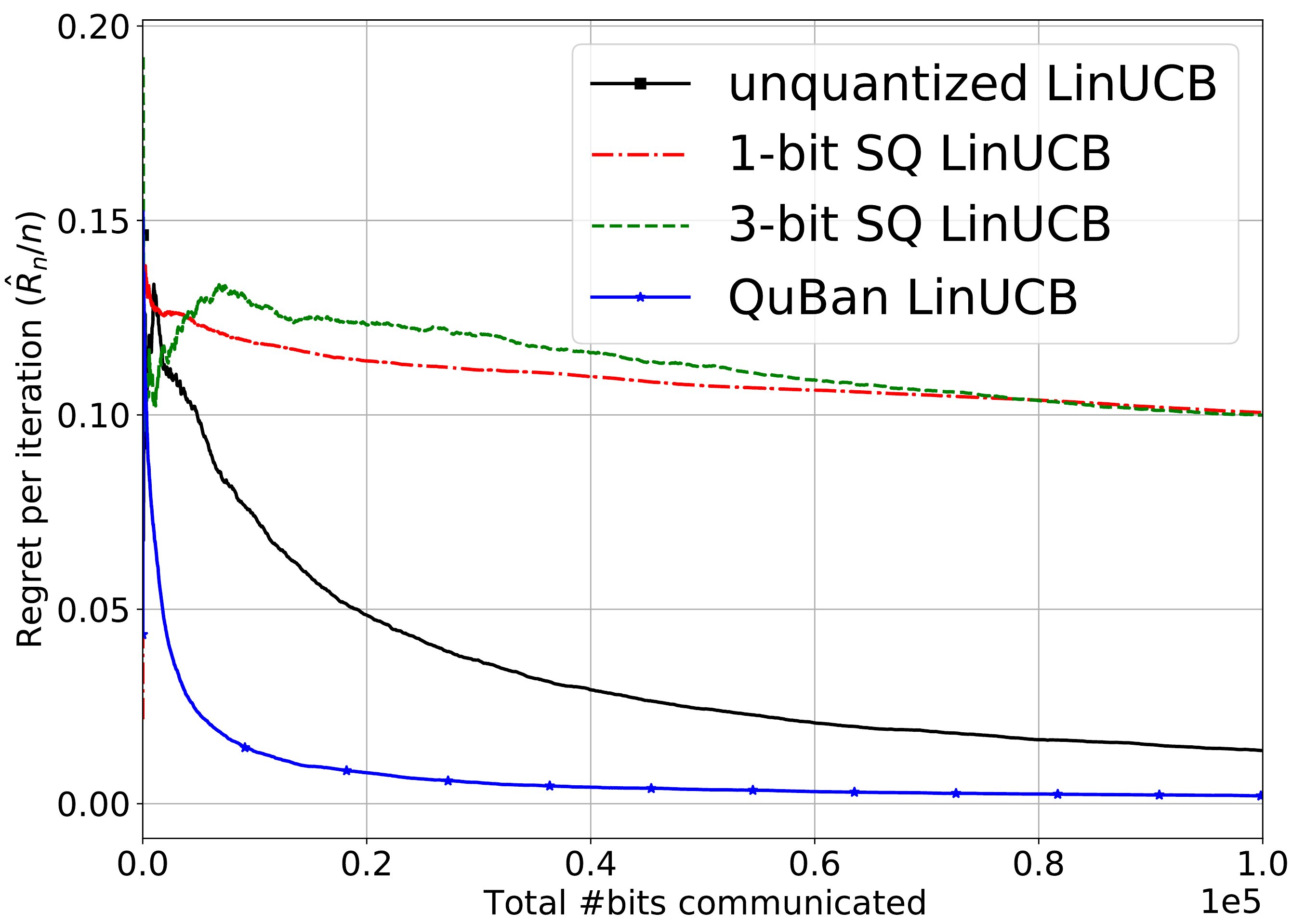}\label{fig:bits_LinUCB}}
  \caption{Total number of bits versus regret per iteration.}\label{fig:bit}
\end{figure*}

We here present our numerical results.

\noindent{\bf Quantization Schemes.} We compare $\qname$ against the baseline schemes described next.\\
\noindent{\em Unquantized.} Rewards are conveyed using the standard 32 bits representation.\\
\noindent{\em r-bit SQ.} We implement r-bit stochastic quantization, 
by using the quantizer described in Section~\ref{sec:model}, with $2^r$ levels uniformly dividing a range $[-\lambda,\lambda]$. \\
\noindent{\em $\qname$.} 
We implement $\qname$ with $\epsilon =X_t=1$.
 
\noindent{\bf MAB Algorithms.} We use quantization on top of:\\ (i) the UCB implementation in \cite[chapter~8]{lattimore2020bandit}. 
The UCB exploration constant is chosen to be $\sigma_q$, an estimate of the standard deviation of the quantized reward distribution.\\
(ii) the $\epsilon$-greedy algorithm in \cite[chapter~6]{lattimore2020bandit},  where $\epsilon_t$ is set to be $\epsilon_t=\min\{1,\frac{C\sigma_qk}{t\Delta_\text{min}^2}\}$.\\
(iii) the LinUCB algorithm for stochastic linear bandits  in \cite[chapter~19]{lattimore2020bandit}.

\begin{figure*}[t!]
  \centering
\subfigure[Setup 1 {(larger $\Delta_i$ values).}]{\includegraphics[width=0.32\linewidth]{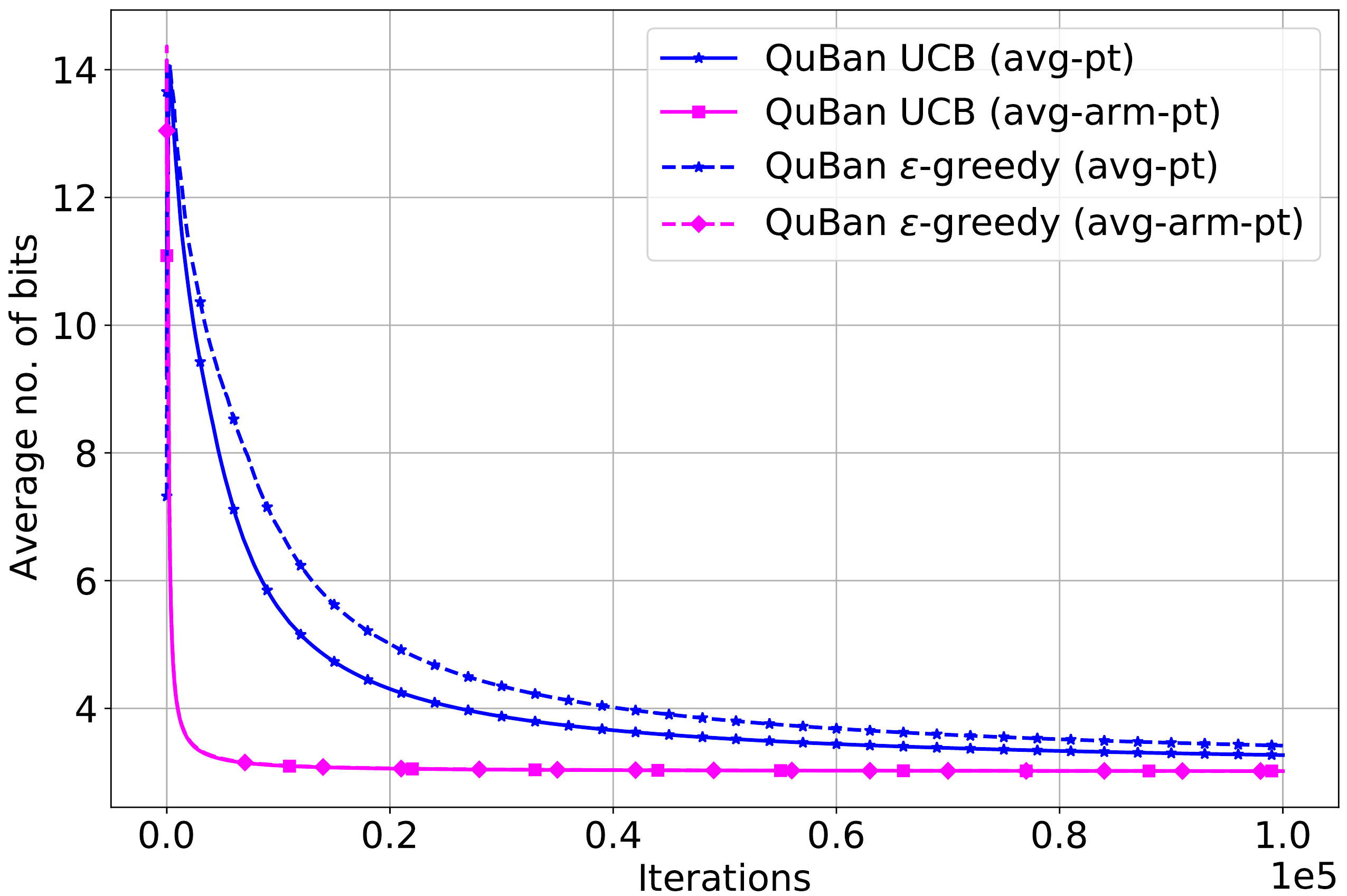}\label{fig:regrets_bits_app}}
  \subfigure[Setup 2 {(smaller $\Delta_i$ values).}]{\includegraphics[width=0.32\linewidth]{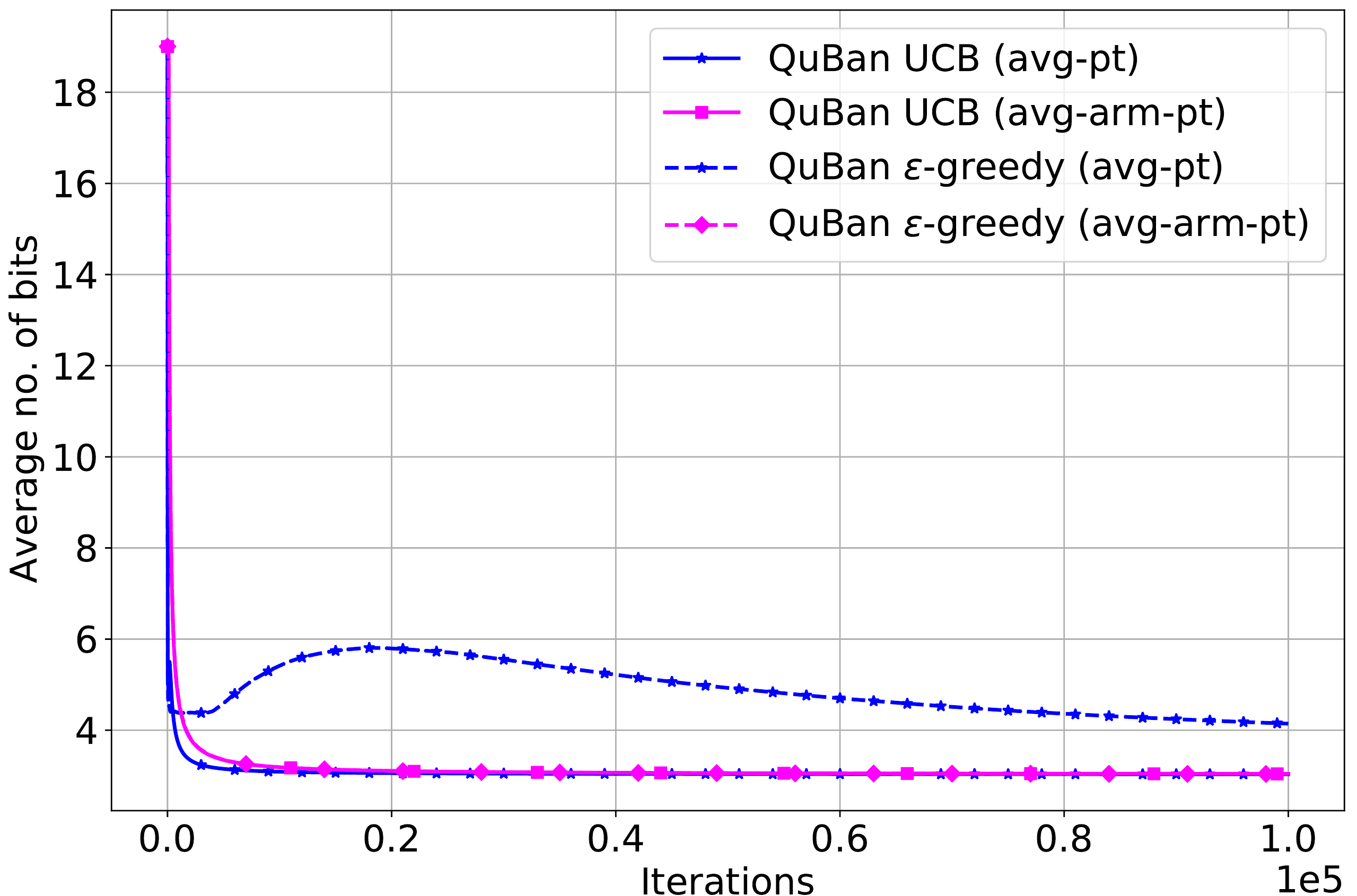}\label{fig:bits_UCB_app}}
  \subfigure[Setup 3 {(linear bandits).}]{\includegraphics[width=0.32\linewidth]{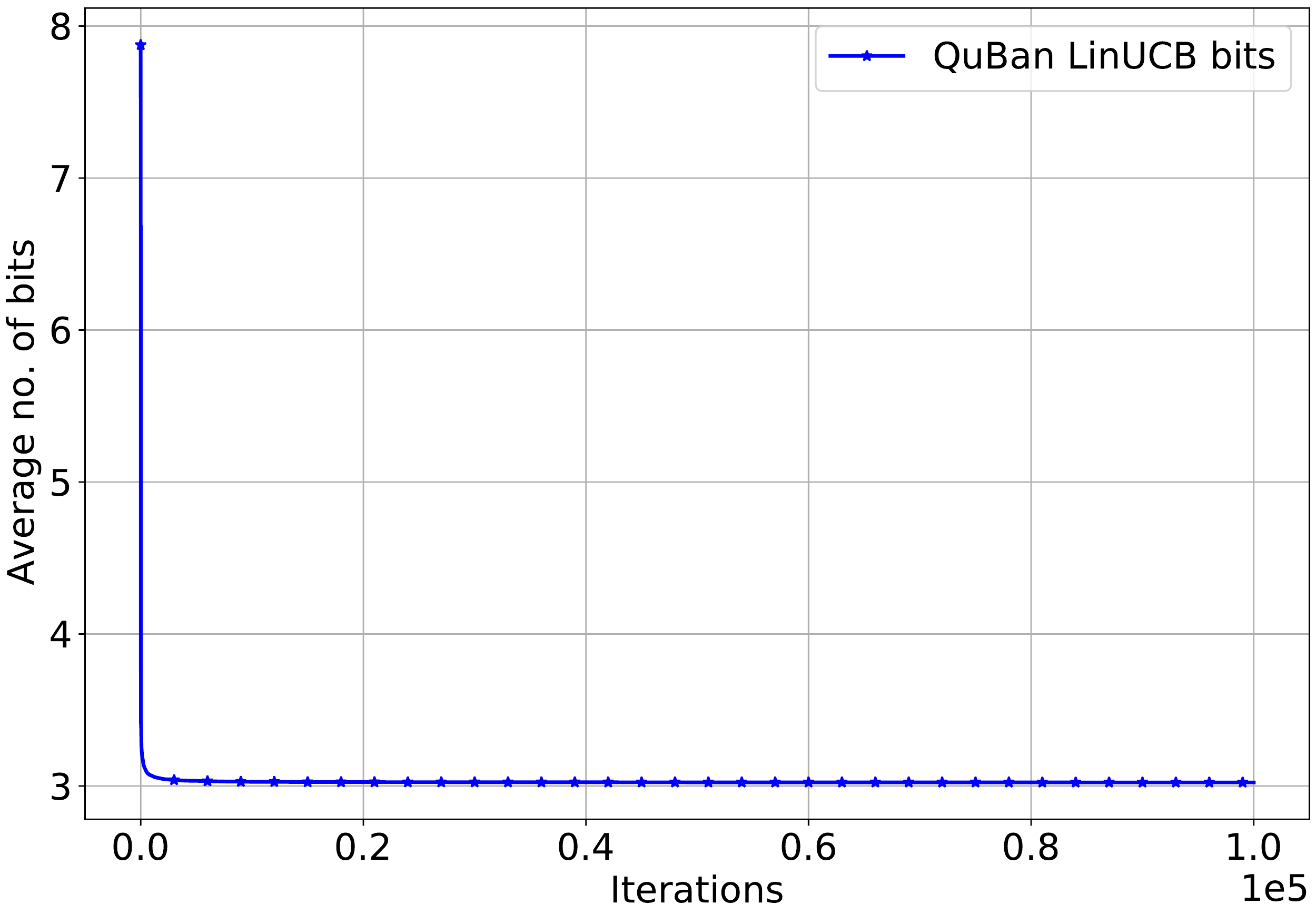}\label{fig:bits_LinUCB_app}}
  \caption{Average number of bits versus iterations.}\label{fig:bit_app}
\end{figure*}

\textbf{\bf MAB Setup.} We simulate three cases. In each case we average over 10 runs of each experiment.\\
$\bullet$ \textbf{Setup 1: (Figs~$3-6$(a)).} We use $k=100, \lambda=100, C=10$, the arms' means are picked from a Gaussian distribution with mean $0$ and standard deviation $10$ and the reward distributions are conditionally Gaussian given the actions $A_t$ with variance $0.1$. The parameter $\sigma_q$ is set to be $0.1$ for $\qname$ and ${200}/{2^r-1}$ for the $r$-bit SQ.\\
$\bullet$ \textbf{Setup 2: (Figs~$3-6$(b))} This differs from the previous only in that the means are picked from a Gaussian distribution {with mean $95$} and standard deviation $1$ (leading to smaller $\Delta_i$).\\
$\bullet$ \textbf{Setup 3: (Figs~$3-6$(c)).}  This is our contextual bandit setup. We use $d=20$ dimensions, $\theta_*$ picked uniformly at random from the surface of a radius $1$ ball centered at the origin, and the noise $\eta_t$ is picked from a Gaussian distribution with zero mean and $0.1$ variance. At each time $t$ we construct the actions set $\mathcal{A}_t$ by sampling $5$ actions uniformly at random from the surface of a radius $0.5$ ball centered at the origin independently of the previously sampled actions. We evaluate the regret and the average number of bits used by $\qname$ as well as the $3$ and $1$ bit stochastic quantizers in the interval $[-10,10]$ (the interval in which we observe the majority of rewards). These quantization schemes are used on top of the LinUCB algorithm. The LinUCB exploration constant is chosen to be $\sigma_q$, where $\sigma_q$ is set to be $0.1$ for $\qname$ and $\frac{20}{2^r-1}$ for the $r$-bit SQ.\\
\begin{figure*}[t!]
  \centering
\subfigure[Setup 1 {(larger $\Delta_i$ values).}]{\includegraphics[width=0.32\linewidth]{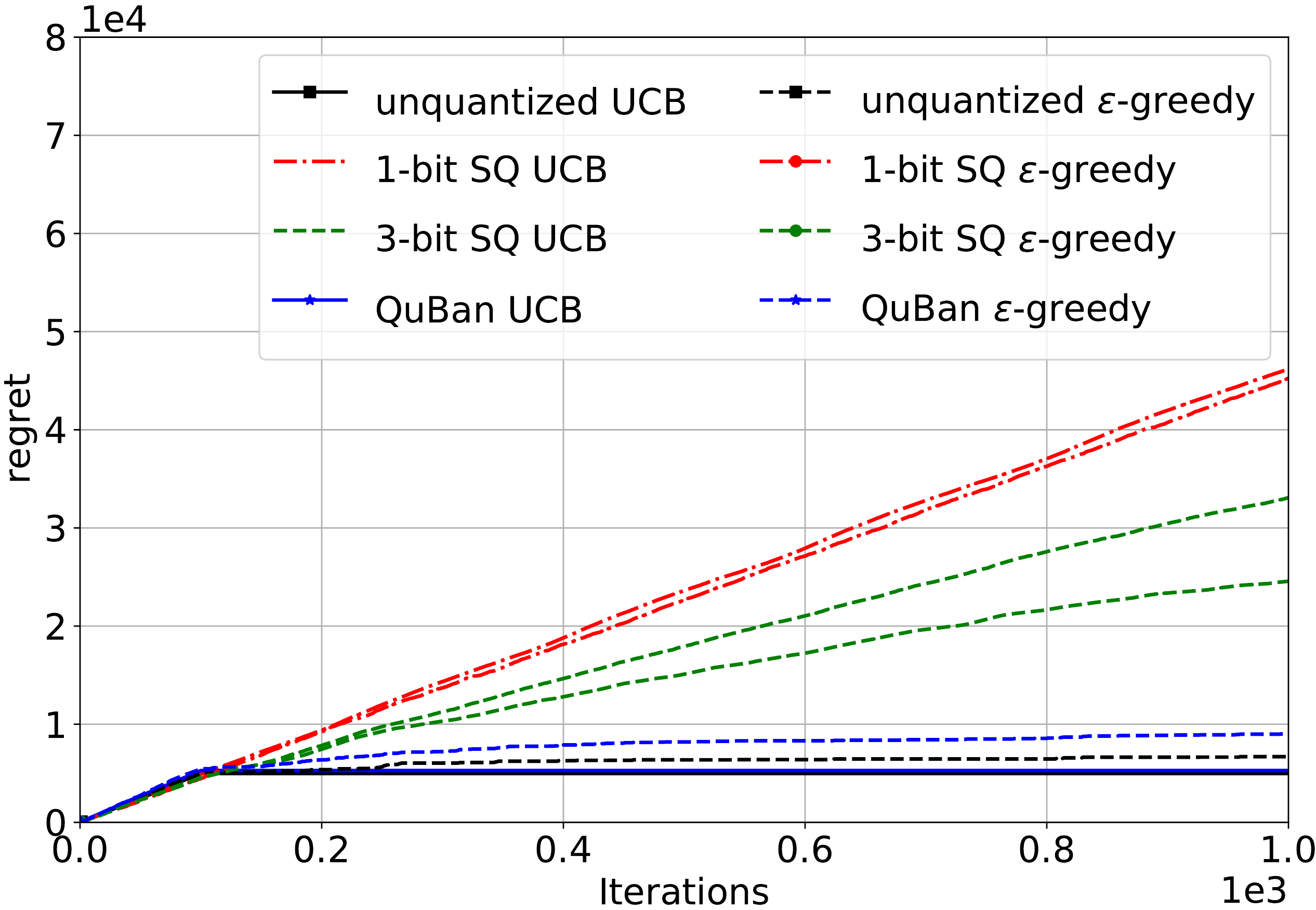}\label{fig:regrets_all_app}}
 \subfigure[Setup 2 {(smaller $\Delta_i$ values).}]{\includegraphics[width=0.32\linewidth]{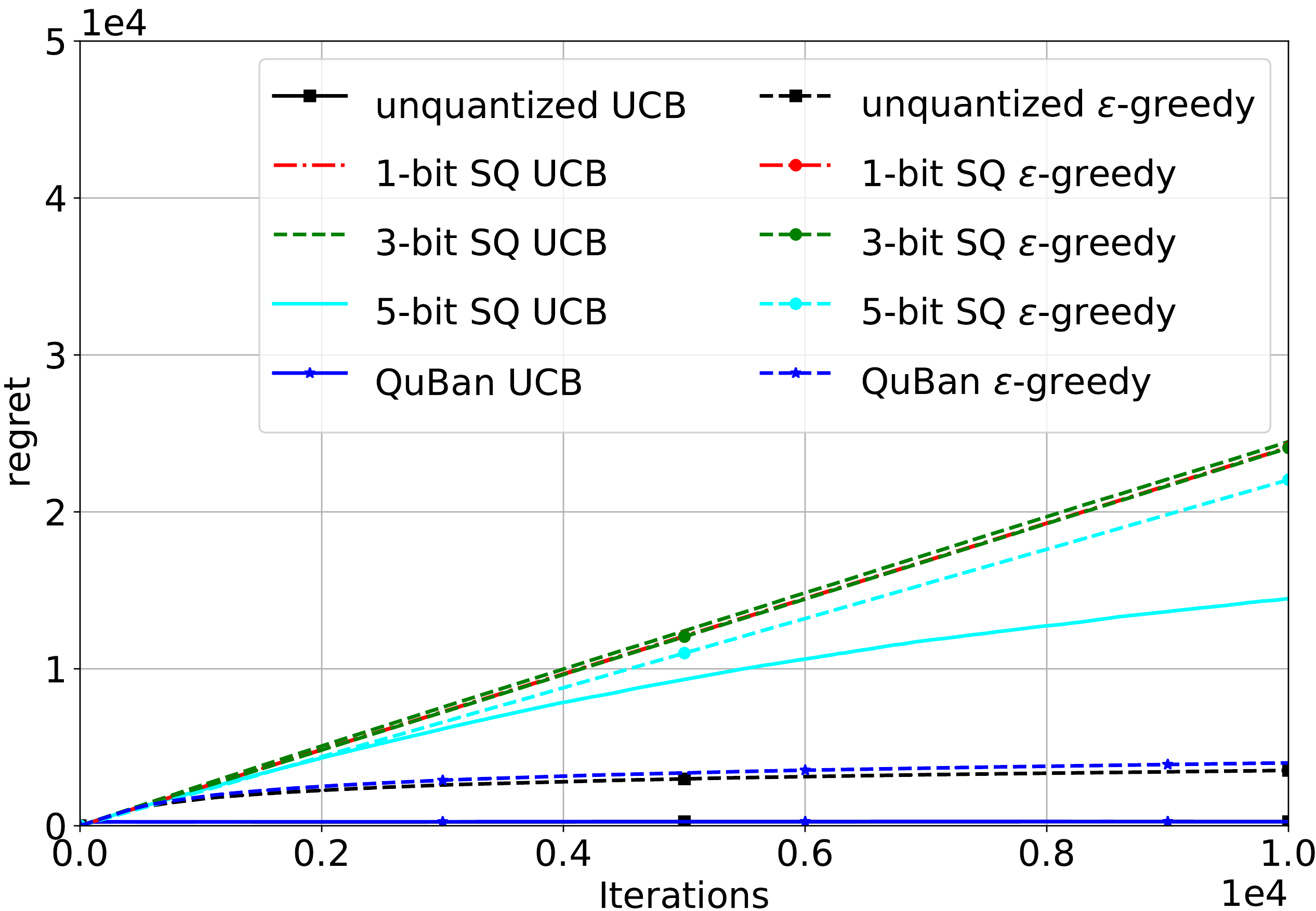}\label{fig:regret_UCB_app}}
  \subfigure[Setup 3 {(linear bandits).}]{\includegraphics[width=0.32\linewidth]{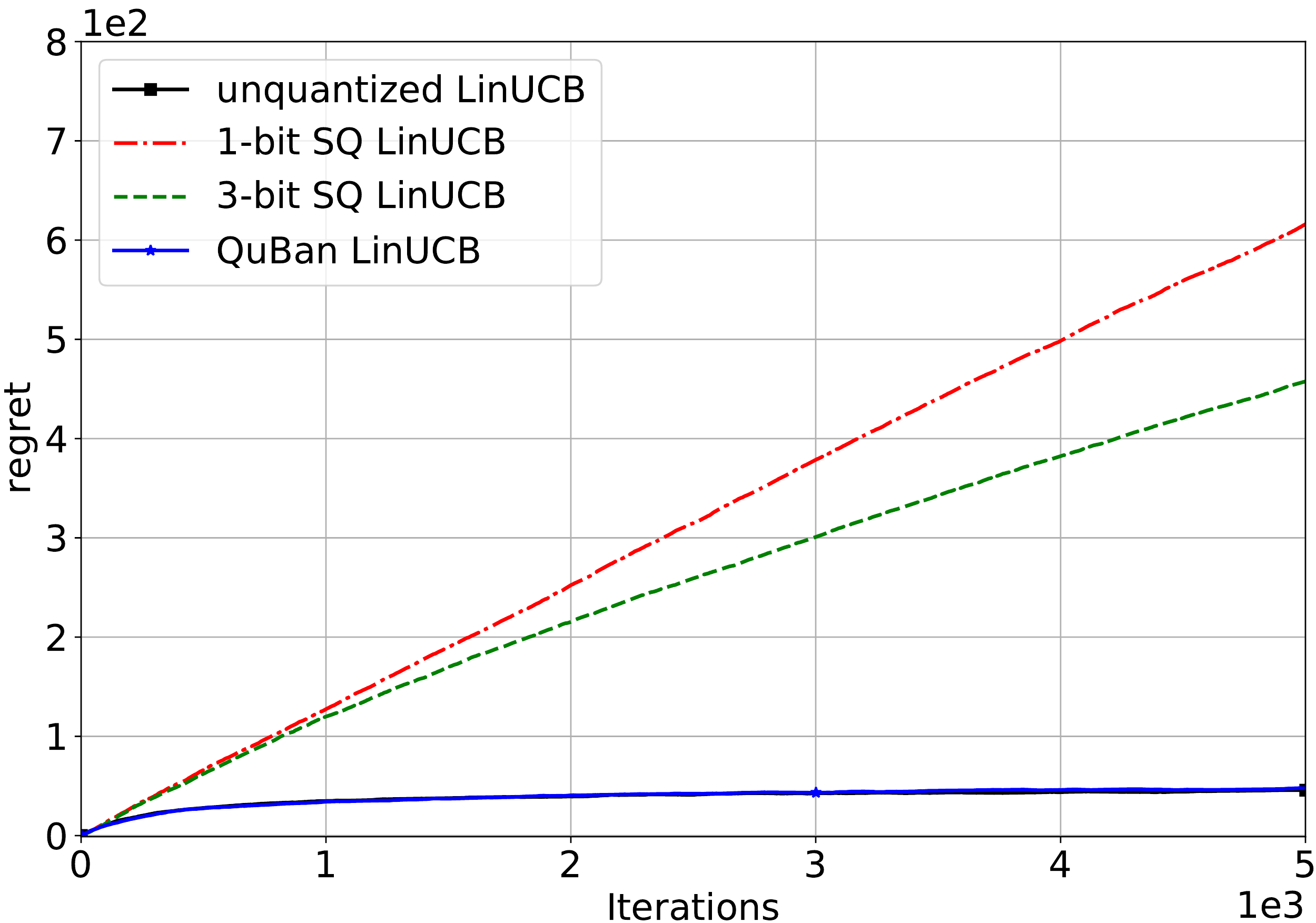}\label{fig:regret_LinUCB_app}}
  \caption{Regret versus number of iterations.}\label{fig:iter_app}
\end{figure*}
\noindent{\bf Results.} 
Fig.~\ref{fig:iter} plots the regret ${R}'_n$ in (\ref{eq:regret}) vs. the number of iterations, Fig.~\ref{fig:bit} plots $\frac{\hat{R}_n}{n}$, the regret per iteration, vs. the total number of bits communicated, Fig.~\ref{fig:iter_app} plots the regret versus number of iterations, and Fig.~\ref{fig:bit_app} plots the average number of bits versus iterations. We find that:\\
$\bullet$ $\qname$ in all three setups offers minimal or no regret increase compared to the unquantized rewards regret  and achieves savings of tens of thousands of bits as compared to unquantized communication.\\
$\bullet$ 1-bit SQ significantly diverges in most cases; 3-bit and 5-bit SQ show better performance yet still not matching $\qname$ with a performance gap that  increases when the arms means are closer ($\Delta_i$ smaller), and hence, more difficult to distinguish.\\
$\bullet$ $\qname$ allows for more than $10$x saving in the number of bits over the unquantized case to achieve the same regret. In all three setups $\qname$ achieves $\mathbb{E}[\Bar{B}(n)] \approx 3$ (see Fig.~\ref{fig:bit_app}).\\
$\bullet$ Both $\qname$ avg-pt  and avg-arm-pt achieve the same regret (they are not distinguishable in Fig.~\ref{fig:iter} and thus we use a common legend), yet avg-arm-pt uses a smaller number of bits when the means of the arms tend to be well separated (Fig.~\ref{fig:regrets_bits}) while avg-pt  uses a smaller number of bits when they tend to be closer together (Fig.~\ref{fig:bits_UCB}). We also observe that the avg-pt tends to perform better for a well-behaved bandit scheme, while the avg-arm-pt performs better when the algorithm picks sub-optimal arms for many iterations (e.g., $\epsilon$-greedy in Fig.~\ref{fig:bits_UCB_app}).\\

 \section{Conclusion and Future Work}\label{sec:concl}
In this paper we provide  a generic framework, $\qname$, to quantize rewards for MAB problems. This framework can be used on top of nearly all the existing and future MAB algorithms, making them attractive for distributed learning applications where communication can become a bottleneck. We have demonstrated that, both in theory and by numerical experiments, $\qname$ can provide very significant savings in terms of communication and barely affects the learning performance. 
We identify several future research directions: (1) How to exploit memory? In the setup we consider,   the remote agents are changing over time, and thus they are essentially memoryless, i.e., a new agent does not know the history information of previous agents. 
(2) How to deal with heavy tailed noise?   (3)  How to convey contexts in the contextual bandit setting if these are not implicitly conveyed? 
Resolving such questions can offer additional benefits for communication-sensitive bandit learning setups.



\bibliographystyle{IEEEtran}
\bibliography{Quant_multiarm}

\newpage
\appendices

\section{Discussion On the Quantization Scheme}\label{app:disc}
In this appendix we discuss some aspects of $\qname$.\\
\underline{\textbf{Description of the algorithm:}}\\
At iteration $t$, $\qname$ centers its quantization around the value $\hat{\mu}(t)$. It then quantizes the normalized reward $\Bar{r}_t={r_t}/{M_t}-\floor{{\hat{\mu}(t)}/{M_t}}$ to one of the two values $\floor{\Bar{r}_t}, \ceil{\Bar{r}_t}$, where $M_t=\epsilon \sigma X_t$, $\epsilon$ is a parameter to control the regret vs number of bits trade-off as will be illustrated later in this section, and $\{X_t\}_{i=1}^n$ are independent samples from a $\frac{1}{4}$-subgaussian distribution satisfying $|X_t|\geq 1$ almost surely, e.g., we can use $X_t=1$ almost surely. If $X_t$ is allowed to take larger values, it will make the quantization coarser with some probability, as will be described next, resulting in less number of bits. For example, if $X_t$ has Gaussian tail, the average number of bits can be improved to $\approx 2.2$ bits. The quantization introduces an error in estimating $\Bar{r}_t$ that is bounded by $1$, which results in error of at most $M_t$ in estimating $r_t=M_t (\Bar{r}_t+\floor{{\hat{\mu}(t)}/{M_t}})$. This quantization is done in a randomized way to convey an unbiased estimate of $r_t$. 
More precisely, the {agent} that pulls the arm at time $t$ does the following operations. If $-3\leq \Bar{r}_t\leq 4$, the agent quantizes $\Bar{r}_t$ to $\ceil{\Bar{r}_t}$ with probability $\Bar{r}_t-\floor(\Bar{r}_t)$, and to $\floor{\Bar{r}_t}$ with probability $\ceil{\Bar{r}_t}-\Bar{r}_t$. It then transmits $3$ bits to distinguish between the $8$ cases: \textbf{(a)} $-2\leq \Bar{r}_t\leq 3$ and $\hat{\Bar{r}}_t=i$), $i\in\{-2,..,3\}$; \textbf{(b)} (a) does not hold and $\Bar{r}_t>0$; \textbf{(c)} (a) does not hold and $\Bar{r}_t<0$.
If (b) holds, the agent transmits one bit to distinguish between the two cases: $\hat{\Bar{r}}_t=4, \Bar{r}_t>4$. If (c) holds, the agent transmits one bit to distinguish between the two cases: $\hat{\Bar{r}}_t=-3, \Bar{r}_t<-3$. If $\Bar{r}_t>4$ or $\Bar{r}_t<-3$ the agent quantizes $\Bar{r}'=|\Bar{r}_t|-|a|$ as follows, where $a=4$ if $\Bar{r}_t>4$, $a=-3$ if $\Bar{r}_t<-3$. The agent conveys the greatest power of $2$ below $\Bar{r}'_t$, call it $2^{I_t}$ (an extra bit is transmitted to separate the case where $|\Bar{r}_t|\leq 1$, and $2^{I_t}$ is assumed to be $0$ in that case). Then, it quantizes $|\Bar{r}_t|-2^{I_t}$ using SQ with levels that are $1$ distance apart in the interval $[0,2^{I_t}]$\footnote{Note that $0\leq |\Bar{r}_t|-2^{I_t}\leq 2^{I_t}$.} (see Fig.~\ref{fig:quant} for an example) (if $|\Bar{r}_t|\leq 1$, SQ is used in the interval $[0,1]$). The agent then transmits $I_t$ with unary coding by transmitting $I_t$ zeros followed by $1$ one\footnote{{Alternative (to unary) coding techniques that can result in a smaller number of bits  are discussed in App.~\ref{app:disc}.}}. This uses $O(\log(\Bar{r}_t))$ bits. The SQ output is transmitted using $O(\log(\Bar{r}_t))$ bits. An estimated value of $r_t$ is obtained from the quantized $\Bar{r}_t$ by a proper shift and scaling. 
We recall that $\frac{\hat{\mu}(t)}{\sigma}$ is believed to be close to $\frac{r_t}{\sigma}$ in the majority of iterations resulting in small values for $\log(\Bar{r}_t)$. Algorithms~\ref{alg:quant} and \ref{alg:quant_reward} describe the operation of $\qname$.

\underline{\textbf{Sending the least power of $2$ below $\Bar{r}_t$:}}\\
For simplicity we consider the case where $\Bar{r}_t\geq 0$. We note that since it is possible for the decoded reward to take any value in the set $\{\floor{\frac{\hat{\mu}}{\sigma}},\floor{\frac{\hat{\mu}}{\sigma}}+1,\floor{\frac{\hat{\mu}}{\sigma}}+2...\}$ (to guarantee the uniform upper bound on $|\hat{r}_t-r_t|$), every value in that set needs to be encoded. A \textit{good} encoding strategy assigns shorter codes to the levels that are close to $\floor{\frac{\hat{\mu}}{\sigma}}$ as they are expected to occur more often. Hence, the best we can hope for is to encode $r_t$ using $O(\log(\frac{r_t}{\sigma}-\floor{\frac{\hat{\mu}}{\sigma}})$ bits as it is quantized to either $\floor{\frac{r_t}{\sigma}}$ or $\ceil{\frac{r_t}{\sigma}}$ and the quantization level at $\floor{\frac{r_t}{\sigma}}$ is encoded using the largest number of bits among the levels in the set $\{\floor{\frac{\hat{\mu}}{\sigma}},\floor{\frac{\hat{\mu}}{\sigma}}+1,\floor{\frac{\hat{\mu}}{\sigma}}+2...,\floor{\frac{r_t}{\sigma}}\}$. As can be seen in Appendix~\ref{app:main_thm}, sending the greatest power of $2$ below $\Bar{r}_t$ then quantizing the difference using SQ gives that $r_t$ is encoded using $O(\log(\frac{r_t}{\sigma}-\floor{\frac{\hat{\mu}}{\sigma}})$ bits. This is achieved since $I_t$ is $O(\log(\frac{r_t}{\sigma}-\floor{\frac{\hat{\mu}}{\sigma}})$ and the SQ uses $2^{I_t}+1$ quantization levels.

\underline{\textbf{Alternatives to unary coding:}}\\
An alternative way to decode $I_t$ is recursively applying our scheme by using unary coding to transmit the largest $I^{(2)}_t$ with $2^{I^{(2)}_t}\leq I_t$ and then encode the difference $I_t-2^{I^{(2)}_t}$ using $\log(1+2^{I^{(2)}_t})$ bits noting that $I_t-2^{I^{(2)}_t}\leq 2^{I^{(2)}_t}$. This results in using $O(\log(\log(\frac{r_t}{\sigma}-\floor{\frac{\hat{\mu}}{\sigma}}))$ bits to encode $I_t$. We keep the unary coding for $I_t$ for simplicity and since it does not dominate the average number of bits.

\underline{\textbf{Preserving regret bounds:}}\\
The main reasons $\qname$ preserves existing regret bounds is that it does not destroy the Markov property (as we prove in Appendix~\ref{app:main_thm}) and it provides that $|\hat{r}_t-r_t|$ is uniformly upper bound. The later property implies that if given $A_t$, $r_t$ is conditionally integrable, sub-exponential, sub-gaussian, or almost surely bounded, then given $A_t$, $\hat{r}_t$ is conditionally integrable, sub-exponential, sub-gaussian, or almost surely bounded respectively. A widely used assumption is that given $A_t$, $r_t$ is conditionally sub-gaussian.

\underline{\textbf{Unknown $\sigma$:}}\\
Throughout the paper, we assume a known upper bound on the noise variance. However, it is not difficult to see that a variance estimate within a constant factor would suffice. Running QuBan with an estimate ${\sigma}'$ that is possibly different from the true $\sigma$ results in a degradation in the regret by a factor of $\max\{1,\frac{{\sigma}'}{\sigma}\}$ and increase in the communication by $2\log(\frac{\sigma}{{\sigma}'})$ bits. 
An optimistic estimate of the noise ${\sigma}'<\sigma$ results in finer quantization, hence, no degradation in the regret at the cost of increasing the communication by $2\log(\frac{\sigma}{{\sigma}'})$ bits.

\section{Proof of Proposition~\ref{prop:1}}\label{app:prop}
\begin{proof}
We  start by proving that $\hat{r}_t$ is an unbiased estimate of $\mu_{A_t}$. If $-3\leq r_t\leq -4$, we have that $\hat{r}_t$ takes the value $\ceil{r_t}$ with probability $r_t-\floor{r_t}$, and the value $\floor{r_t}$ with probability $\ceil{r_t}-r_t$. Hence, $\mathbb{E}[\hat{r}_t|r_t]=r_t$. For all the other cases we have that
\begin{align}
\mathbb{E}[\hat{r}_t|r_t]&= \mathbb{E}[M_t( s_te_t^{(q)}+\lfloor \frac{\hat{\mu}(t)}{M_t}\rfloor+s_t\ell_t)|r_t]\nonumber \\
&= \mathbb{E}[M_t\mathbb{E}[s_te_t^{(q)}+\lfloor \frac{\hat{\mu}(t)}{M_t}\rfloor+s_t\ell_t|r_t,\hat{\mu}(t),M_t]|r_t]\nonumber \\
&\stackrel{(i)}{=}  \mathbb{E}[M_t(\frac{r_t}{M_t}-(\lfloor \frac{\hat{\mu}(t)}{M_t}\rfloor+s\ell_t)+\lfloor \frac{\hat{\mu}(t)}{M_t}\rfloor+s\ell_t)|r_t]\nonumber \\
&= r_t,
\end{align}
where $(i)$ follows from the fact that the stochastic quantization (SQ) that we use gives an unbiased estimate of the input. Hence, in all cases we have that
\begin{align}
    \mathbb{E}[\hat{r}_t|A_t] = \mathbb{E}[\mathbb{E}[\hat{r}_t|r_t,A_t]|A_t]=\mathbb{E}[\mathbb{E}[\hat{r}_t|r_t]|A_t]= \mathbb{E}[r_t|A_t]=\mu_{A_t}
\end{align}

The bound on $|r_t-\hat{r}_t|$ follows from the fact that the distance between the quantization levels for which we use the randomized quantization is $1$, hence, in all cases we have that
\begin{equation}
    1\geq |s_te_t^{(q)}-(\frac{r_t}{M_t}-\lfloor {\frac{\hat{\mu}(t)}{M_t}}\rfloor-s_t\ell_t)|=\frac{|\hat{r}_t-r_t|}{M_t}.
\end{equation}
We note that this implies
\begin{align}
    \mathbb{E}[|\hat{r}_t-\mu_{A_t}|^2|A_t]&=\mathbb{E}[|\hat{r}_t-r_t+r_t-\mu_{A_t}|^2|A_t]\nonumber\\
    &=\mathbb{E}[|\hat{r}_t-r_t|^2|A_t]+\mathbb{E}[|r_t-\mu_{A_t}|^2|A_t]+2\mathbb{E}[(r_t-\mu_{A_t})(\hat{r}_t-r_t)|A_t]\nonumber\\
    &\leq (1+\epsilon^2)\sigma^2+2\mathbb{E}[(r_t-\mu_{A_t})\mathbb{E}[(\hat{r}_t-r_t)|A_t,r_t]|A_t]\nonumber\\
    &=(1+\epsilon^2)\sigma^2.
\end{align}
To see that conditioned on $A_t$, $\hat{r}_t$ is conditionally independent on the history $A_1,\hat{r}_1,...,A_{t-1},\hat{r}_{t-1}$, we notice that since we replace $\frac{\hat{\mu}(t)}{M_t}$ by an integer, $\floor{\frac{\hat{\mu}(t)}{M_t}}$ and since the distance between the quantization levels is $1$, we have that the two nearest quantization levels to $\frac{r_t}{M_t}$ are at $\floor{\frac{r_t}{M_t}},\ceil{\frac{r_t}{M_t}}$. Hence, conditioned on $M_t$, $\hat{r}_t$ takes the value $M_t \ceil{\frac{r_t}{M_t}}$ with probability $\frac{r_t}{M_t}-\floor{\frac{r_t}{M_t}}$, and the value $M_t \floor{\frac{r_t}{M_t}}$ with probability $\ceil{\frac{r_t}{M_t}}-\frac{r_t}{M_t}$. This shows that despite the fact that the encoding of $\hat{r}_t$ is a function of $r_1,...,r_t$, the value of $\hat{r}_t$ is a function of $r_t$ only, since $M_t$ is generated independently of the history. As a result, given $A_t$, $\hat{r}_t$ is conditionally independent on the history $A_1,\hat{r}_1,...,A_{t-1},\hat{r}_{t-1}$.

The fact that $\hat{r}_t$ is subgaussian can be proven by Cauchy-Schwarz
\begin{align}
    \mathbb{E}[e^{\lambda (\hat{r}_t-\mu_{A_t})}|A_t]&=\mathbb{E}[e^{\lambda (\hat{r}_t-r_t+r_t-\mu_{A_t})}|A_t]\nonumber \\
    &\leq {\mathbb{E}[e^{p\lambda (\hat{r}_t-r_t)}|A_t]^{\frac{1}{p}}\mathbb{E}[e^{(1-p)\lambda (r_t-\mu_{A_t})}|A_t]^{\frac{1}{1-p}}}\nonumber \\
    &\leq e^{\lambda^2 \frac{\sigma^2(1+\frac{\epsilon}{2})^2}{2}},
\end{align}
where $p=1+\frac{2}{\epsilon}$. To bound the expected regret after quantization we observe that $R_n=\sum_{t=1}^n\mathbb{E}(\mu^*_t-r_t)=\sum_{t=1}^n\mathbb{E}(\mu^*_t-\hat{r}_t)=(1+\frac{\epsilon}{2})\sum_{t=1}^n\mathbb{E}(\frac{\mu^*_t-\hat{r}_t}{1+\frac{\epsilon}{2}})$. We have that $\frac{\hat{r}_t}{(1+\frac{\epsilon}{2})}$ is $\sigma^2$-subgaussian. Applying the bandit algorithm using $\frac{\hat{r}_t}{(1+\frac{\epsilon}{2})}$ results in $\sum_{t=1}^n\mathbb{E}(\frac{\mu^*_t-\hat{r}_t}{(1+\frac{\epsilon}{2})})\leq R^U_n(\{\Delta_i/(1+\frac{\epsilon}{2})\})$, hence
\begin{equation}
    R_n\leq (1+\frac{\epsilon}{2})R^U_n(\{\Delta_i/(1+\frac{\epsilon}{2})\}).
\end{equation}
\end{proof}
\section{Proof of Theorem~\ref{main_thm}}\label{app:main_thm}
\begin{proof}
We have that $B_t$ can be bounded as
\begin{align}
    B_t&\leq 3+\mathbf{1}[\frac{r_t}{M_t}-\floor{\frac{\hat{\mu}(t)}{M_t}}>3]+\mathbf{1}[\floor{\frac{\hat{\mu}(t)}{M_t}}-\frac{r_t}{M_t}>2]\nonumber \\
    &\quad +2(\mathbf{1}[\frac{r_t}{M_t}-\floor{\frac{\hat{\mu}(t)}{M_t}}>4]\ceil{\log(\frac{r_t}{M_t}-\floor{\frac{\hat{\mu}(t)}{M_t}}-3)})\nonumber \\
    &\quad +2(\mathbf{1}[\floor{\frac{\hat{\mu}(t)}{M_t}}-\frac{r_t}{M_t}>3]\ceil{\log(\floor{\frac{\hat{\mu}(t)}{M_t}}-\frac{r_t}{M_t}-2)})\nonumber \\
    &\leq 3+\mathbf{1}[|\frac{r_t}{M_t}-{\frac{\hat{\mu}(t)}{M_t}}|>2]+2(\mathbf{1}[|\frac{r_t}{M_t}-{\frac{\hat{\mu}(t)}{M_t}}|>3])\nonumber \\
    &\quad +2(\mathbf{1}[|\frac{r_t}{M_t}-{\frac{\hat{\mu}(t)}{M_t}}|>3]{\log(|\frac{r_t}{M_t}-{\frac{\hat{\mu}(t)}{M_t}}|-2)}).
\end{align}
Hence for each $\delta>0$, we have
\begin{align}\label{eq:noavg}
    B_t &\leq 3+\mathbf{1}[|\frac{r_t-\mu_{A_t}}{\sigma}|>2(1-\delta)]+\mathbf{1}[|\frac{\mu_{A_t}-\hat{\mu}(t)}{\sigma}|>2\delta]\nonumber \\
    &\quad +2(\mathbf{1}[|\frac{r_t-\mu_{A_t}}{\sigma}|>3(1-\delta)]+\mathbf{1}[|\frac{\mu_{A_t}-\hat{\mu}(t)}{\sigma}|>3\delta])\nonumber \\
    &\quad +2(\mathbf{1}[|\frac{r_t-\mu_{A_t}}{\sigma}|>3]){\log(|\frac{r_t-\hat{\mu}(t)}{\sigma}|-2)}.
\end{align}
Taking the expectation of both sides, we get that
\begin{align}
    \mathbb{E}[B_t] &\leq 3+\mathbb{P}[|\frac{r_t-\mu_{A_t}}{\sigma}|>2(1-\delta)]+\mathbb{P}[|\frac{\mu_{A_t}-\hat{\mu}(t)}{\sigma}|>2\delta]\nonumber \\
    &\quad +2(\mathbb{P}[|\frac{r_t-\mu_{A_t}}{\sigma}|>3(1-\delta)]+\mathbb{P}[|\frac{\mu_{A_t}-\hat{\mu}(t)}{\sigma}|>3\delta])\nonumber \\
    &\quad +2\mathbb{E}[(\mathbf{1}[|\frac{r_t-\mu_{A_t}}{\sigma}|>3]){\log(|\frac{r_t-\hat{\mu}(t)}{\sigma}|-2)}].
\end{align}

Hence, there are universal constants $C$, $C'$ such that 
\begin{align}\label{eq:avged}
    \mathbb{E}[B_t]&\leq 3.32+C'\mathbb{E}[|\frac{\mu_{A_t}-\hat{\mu}(t)}{\sigma}|]+2\mathbb{E}[\mathbf{1}[|\frac{r_t}{M_t}-{\frac{\hat{\mu}(t)}{M_t}}|>3]{(|\frac{r_t}{M_t}-{\frac{\hat{\mu}(t)}{M_t}}|-3)}]\nonumber \\
    &\leq 3.32+C'\mathbb{E}[|\frac{\mu_{A_t}-\hat{\mu}(t)}{\sigma}|]+2\mathbb{E}[\mathbf{1}[|\frac{r_t-\mu_{A_t}}{\sigma}|>3(1-\delta)]{||\frac{r_t-\mu_{A_t}}{\sigma}|-3|}]
\nonumber \\
&\quad +2\mathbb{E}[\mathbf{1}[|\frac{\mu_{A_t}-\hat{\mu}(t)}{\sigma}|>3\delta]{||\frac{r_t-\mu_{A_t}}{\sigma}|-3|}]+2\mathbb{E}[{|\frac{\mu_{A_t}-\hat{\mu}(t)}{\sigma}|}]\nonumber \\
    &\leq 3.32+(C'+2)\mathbb{E}[|\frac{\mu_{A_t}-\hat{\mu}(t)}{\sigma}|]+2\sum_{i=3}^{\infty}|i(1-\delta)-3|\mathbb{P}[|\frac{\mu_{A_t}-\hat{\mu}(t)}{\sigma}|>i(1-\delta)] \nonumber \\
    &\quad +2\mathbb{E}[\mathbf{1}[|\frac{\mu_{A_t}-\hat{\mu}(t)}{\sigma}|>3\delta]]\mathbb{E}[{||\frac{r_t-\mu_{A_t}}{\sigma}|-3|}]+2\mathbb{E}[{|\frac{\mu_{A_t}-\hat{\mu}(t)}{\sigma}|}]\nonumber \\
    &\leq 3.4+C\mathbb{E}[|\frac{\mu_{A_t}-\hat{\mu}(t)}{\sigma}|]
\end{align}

From \eqref{eq:noavg}, $\mathbb{E}[|r_t-\mu_{A_t}|^2|A_t]\leq \sigma^2$, Markov property and the strong law of large numbers for martingales, we also have that there is a universal constant $C$ such that
\begin{align}\label{eq:a.s}
    \lim_{n\to \infty}\frac{1}{n}\sum_{t=1}^nB_t&\stackrel{}{\leq} 3.4+\lim_{n\to \infty}\frac{C}{n}{\sum_{t=1}^n |\frac{\mu_{A_t}-\hat{\mu}(t)}{\sigma}|} \text{\quad almost surely}.
\end{align}
It then remains to analyze $|\mu_{A_t}-\hat{\mu}(t)|$ for the three proposed choices of $\hat{\mu}(t)$. \\

\underline{\textbf{$\bullet$ avg-pt ($\hat{\mu}(t)=\frac{1}{t-1}\sum_{j=1}^{t-1}\hat{r}_j$):}}\\
We have that for $t>1$
\begin{align}
    \frac{|\mu_{A_t}- \hat{\mu}(t)|}{\sigma}&\leq \frac{|\mu_{A_t}- \mu^*|}{\sigma}+\frac{|\mu^*- \hat{\mu}(t)|}{\sigma}\nonumber \\
    &=\frac{\Delta_{A_t}}{\sigma}+|\frac{\sum_{j=1}^{t-1}\mu^*-\mu_{A_j}+\mu_{A_j}-\hat{r}_j}{(t-1)\sigma}|\nonumber \\
    &\leq \frac{\Delta_{A_t}}{\sigma}+|\frac{\sum_{j=1}^{t-1}\mu^*-\mu_{A_j}}{(t-1)\sigma}|+|\frac{\sum_{j=1}^{t-1}\mu_{A_j}-\hat{r}_j}{(t-1)\sigma}|\nonumber \\
    &=\frac{\Delta_{A_t}}{\sigma}+\frac{\sum_{i=1}^k\Delta_{i}T_i(t-1)}{(t-1)\sigma}+|\frac{\sum_{j=1}^{t-1}\mu_{A_j}-\hat{r}_j}{(t-1)\sigma}|.
\end{align}

For $t=1$ we have
\begin{align}
    \frac{|\mu_{A_t}- \hat{\mu}(t)|}{\sigma}&\leq \frac{|\mu_{A_t}- \mu^*|}{\sigma}+\frac{|\mu^*- \hat{\mu}(t)|}{\sigma}\nonumber \\
    &=\frac{\Delta_{A_1}}{\sigma}+\frac{|\mu^*|}{\sigma}.
\end{align}

We then have that
\begin{align}\label{eq:mean_diff}
    \frac{1}{n}\sum_{t=1}^n \log(1+&|\frac{\mu_{A_t}-\hat{\mu}(t)}{\sigma}|)\leq \frac{\log(1+\frac{|\mu^*|}{\sigma})}{n\sigma}+\frac{1}{n}\sum_{t=1}^n \log(1+\frac{\Delta_{A_t}}{\sigma})\nonumber \\
    &+\frac{1}{n}\sum_{t=2}^n\log(1+\frac{\sum_{i=1}^k\Delta_{i}T_i(t-1)}{(t-1)\sigma})+\log(1+|\frac{\sum_{j=1}^{t-1}\mu_{A_j}-\hat{r}_j}{(t-1)\sigma}|)\nonumber \\
    \leq& \frac{\log(1+\frac{|\mu^*|}{\sigma})}{n\sigma}+\frac{1}{n}\left( \frac{\sum_{i=1}^k\Delta_{i}T_i(n)}{\sigma}+\sum_{t=1}^{n-1}\frac{\sum_{i=1}^k\Delta_{i}T_i(t)}{t\sigma}+|\frac{\sum_{j=1}^{t}\mu_{A_j}-\hat{r}_j}{t\sigma}|\right).
\end{align}
We have that since $\mathbb{E}[|r_t-\mu_{A_t}|^2|A_t]\leq \sigma^2$, and Markov property, then by the strong law of large numbers for martingales $\lim_{t\to \infty}\frac{\sum_{j=1}^{t-1}\mu_{A_j}-\hat{r}_j}{(t-1)\sigma}=0$ almost surely. We then have that if the limit of average regret is $0$ almost surely (or in probability), then from \eqref{eq:a.s} and \eqref{eq:mean_diff} we get that 
\begin{equation}
    \lim_{n\to \infty}\frac{1}{n}\sum_{t=1}^nB_t{\leq} 3.4 \text{\quad almost surely (or in probability)}.
\end{equation}
By observing that we can generate a long sequence of rewards from each arm before the process starts and since $\mathbb{E}[|r_t-\mu_{A_t}|^2|A_t]\leq \sigma^2$, then by the triangle inequality we have that
\begin{align}\label{eq:int_bnd}
    \frac{1}{n}\sum_{t=1}^n\mathbb{E}[|\frac{\sum_{j=1}^{t-1}\mu_{A_j}-\hat{r}_j}{(t-1)\sigma}|]&\stackrel{(i)}{\leq}\frac{2}{n}\sum_{t=1}^n \frac{1}{\sqrt{t}}\nonumber \\
    &=\frac{2}{n}\sum_{t=1}^n \frac{1}{\sqrt{t}}\nonumber \\
    &\leq \frac{2}{n}(1+\int_{t=1}^n \frac{1}{\sqrt{t}}dt)\nonumber \\
    &\leq \frac{4}{\sqrt{n}},
\end{align}
where $(i)$ follows from the fact that $\mu_{A_j}-\hat{r}_j, \mu_{A_i}-\hat{r}_i$ are uncorrelated for all $i< j$ since
\begin{equation}
    \mathbb{E}[(\mu_{A_j}-\hat{r}_j)(\mu_{A_i}-\hat{r}_i)]=\mathbb{E}[\mathbb{E}[(\mu_{A_j}-\hat{r}_j)(\mu_{A_i}-\hat{r}_i)|A_j,A_i,\hat{r}_i]]=0.
\end{equation}
We conclude that there is a universal constant $C$ such that
\begin{equation}
    \hat{B}(n)\leq 3.4+(C/n)\left({1+ \log(1+{|\mu^*|}/{\sigma})+{R_n}/{\sigma}+\sum_{t=1}^{n-1}{R_t}/{(\sigma t)}}\right)+{C}/{\sqrt{n}}
\end{equation}
\underline{\textbf{$\bullet$ avg-arm-pt ($\hat{\mu}(t)=\hat{\mu}_{A_t}(t-1)$):}}\\
We have that for $T_{A_t}(t-1)>0$
\begin{align}
    \frac{|\mu_{A_t}- \hat{\mu}(t)|}{\sigma}&=|\frac{\sum_{j=1}^{t-1}(\mu_{A_t}-\hat{r}_j)\mathbf{1}(A_j=A_t)}{T_{A_{t}}(t-1)\sigma}|.\label{eq:av_arm_0}
\end{align}
For $T_{A_t}(t-1)=0$, we have that $\hat{\mu}(t)=0$. Then
\begin{align}
    \frac{1}{n}\sum_{t=1}^n\mathbb{E}[\log(1+\frac{|\mu_{A_t}- \hat{\mu}(t)|}{\sigma})]&\stackrel{(i)}{\leq} \frac{1}{n}{\sum_{i=1}^k \log(1+\frac{|\mu_i|}{\sigma})}+\frac{2}{n}\sum_{t=1}^n \frac{1}{\sqrt{t}}\nonumber \\
    &\stackrel{(ii)}{\leq} \frac{1}{n}{\sum_{i=1}^k \log(1+\frac{|\mu_i|}{\sigma})}+\frac{4}{\sqrt{n}}
\end{align}
where $(ii)$ is as in \eqref{eq:int_bnd}, and $(i)$ can be seen by observing that we can generate a long sequence of rewards from each arm before the process starts, from the fact that $\hat{r}_j-\mu_{A_j}, \hat{r}_i-\mu_{A_i}$ are uncorrelated for all $i\neq j$ and since $\mathbb{E}[|r_t-\mu_{A_t}|^2|A_t]\leq \sigma^2$.

We conclude that there is a universal constant $C$ such that
\begin{equation}
    \hat{B}(n)\leq 3.4+\frac{C}{n}{\sum_{i=1}^k \log(1+\frac{|\mu_i|}{\sigma})}+\frac{C}{\sqrt{n}}.
\end{equation}
The fact that $\lim_{n\to \infty}\frac{1}{n}\sum_{t=1}^nB_t{\leq} 3.4 \text{\ almost surely}$, can be seen using the strong law of large numbers by observing that we can generate a long sequence of rewards from each arm before the process starts, the number of arms is finite, and if $\lim_{n\to \infty}T_i(n)< \infty$ then the contribution of arm $i$ in the number of bits decays to zero almost surely as $n\to \infty$.

\underline{\textbf{$\bullet$ stochastic linear bandits ($\hat{\mu}(t)=\langle \theta_t,A_t \rangle$):}}\\
The results follow directly from \eqref{eq:noavg}, \eqref{eq:avged}, \eqref{eq:a.s} and choice of $\hat{\mu}(t)$.

For the case where $\epsilon\neq 1$, it is easy to see that for small values of $\epsilon$, the number of transmitted bits increases by $2\log(\frac{1}{\epsilon})$ bits. This can be further decreased to $\log(\frac{1}{\epsilon})+\log(\log(\frac{1}{\epsilon}))$ bits using the encoding in App.~\ref{app:disc}. 
\end{proof}
\section{Proof of the High Probability Bound}\label{app:high_prob}
From App.\ref{app:disc}, we have that
\begin{align}\label{eq:high prob}
    B_t&\leq 3+\mathbf{1}[\frac{r_t}{M_t}-\floor{\frac{\hat{\mu}(t)}{M_t}}>3]+\mathbf{1}[\floor{\frac{\hat{\mu}(t)}{M_t}}-\frac{r_t}{M_t}>2]\nonumber \\
    &\quad +\mathbf{1}[\frac{r_t}{M_t}-\floor{\frac{\hat{\mu}(t)}{M_t}}>4]\left(\ceil{\log(\frac{r_t}{M_t}-\floor{\frac{\hat{\mu}(t)}{M_t}}-3)}+\ceil{\log(\log(\frac{r_t}{M_t}-\floor{\frac{\hat{\mu}(t)}{M_t}}-3))}\right)\nonumber \\
    &\quad +\mathbf{1}[\floor{\frac{\hat{\mu}(t)}{M_t}}-\frac{r_t}{M_t}>3]\left(\ceil{\log(\floor{\frac{\hat{\mu}(t)}{M_t}}-\frac{r_t}{M_t}-2)}+\ceil{\log(\log(\floor{\frac{\hat{\mu}(t)}{M_t}}-\frac{r_t}{M_t}-2))}\right)\nonumber \\
    &\leq 4+{\log({\frac{\hat{\mu}(t)}{\sigma}}-\frac{r_t}{\sigma}-2)}+{\log(\log({\frac{\hat{\mu}(t)}{\sigma}}-\frac{r_t}{\sigma}-2))}.
\end{align}
Let the event $G$ be that $\forall t\in \{1,...,n\}: |r_t-\mu_{A_t}|\leq \sigma \sqrt{4\log(n)}$. From the subgaussian assumption and applying the union bound we have that
\begin{equation}
    \mathbb{P}[G]>1-\sum_{t=1}^n e^{-2\log(n)}.
\end{equation}
We have that if $G$ holds then for $t$ with $T_t(A_t)>0$, we have that $|\hat{\mu}(t)-\mu_{A_t}|\leq \sigma, |r_t-\mu_{A_t}|\leq \sigma$. Hence, $|\hat{\mu}(t)-r_t|\leq 2\sigma$. Substituting in~\eqref{eq:high prob}, we get the desired result.
\section{Proof of Lower Bound (Theorem~\ref{thm:lower})}\label{app:lower}
\begin{proof}
To simplify notations, we omit the time index $t$ and only mention it when it is necessary. Normalizing the rewards by $\sigma$, it suffices to consider the case where $\sigma=1$. We will also consider quantization schemes where the set of quantization boundaries is deterministic and fixed over time. The case where the set of quantization boundaries is random follows by similar arguments, while the case where the set of quantization boundaries changes over time follows by observing that the proved properties will hold except for a sublinear number of iterations. We start by showing that to satisfy $(i)$, it suffices to consider unbiased quantization schemes, i.e., quantization schemes that satisfy:\\
$\noindent (iii)$ $\mathbb{E}[\hat{r}|r]=r$.

Let $P, P'$ denote reward distributions with means $\mu_1, \mu_2$ respectively. We have that, for any given algorithm, either: \textbf{Case 1}: $\forall P,P'$ with $\mu_1\neq \mu_2$, we have that $\mathbb{E}_{P}[\hat{r}]\neq \mathbb{E}_{P'}[\hat{r}]$; or \textbf{Case 2}: $\exists P,P'$ with $\mu_1\neq \mu_2$, and $\mathbb{E}_{P}[\hat{r}]= \mathbb{E}_{P'}[\hat{r}]$.

For any algorithm that satisfies the property of \textbf{Case 1}, we will show that we can create another algorithm that satisfies (iii) and achieves the same performance as the original algorithm. In particular, we will prove that, for any algorithm satisfying Case 1, the function that maps $x_i$ to $\mathbb{E}_{P_{x_i}}[x_i]$ has to be of the form $\mathbb{E}_{P_{x_i}}[x_i]=c_1x_i+c_2$ for some constants $c_1,c_2$, and thus, by a proper shift and scaling, the quantization algorithm can be modified to be unbiased without affecting the number of transmitted bits or the performance. 
To do so, we first construct distributions  $P$ and $P'$ as follows.  
Consider the set of distributions $\{P_x\}_{x\in\mathbb{R}}$, where $P_x$ represents the random variable that takes the value $x$ almost surely. Let $\{x_i,p_i, p'_i\}_{i=1}^3$ be real values such that $x_i\neq x_j\forall i\neq j, \sum_{i=1}^3p_i=\sum_{i=1}^3p'_i=1$ and $p_i, p'_i\geq 0 \forall i\in \{1,2,3\}$. We design $P$ to be the distribution of a random variable that takes the value $x_i$ with probability $p_i$, and $P'$ be the distribution of a random variable that takes the value $x_i$ with probability $p'_i$. 

Note that, for the condition of Case 1 to be satisfied, we want to have that 
$\mathbb{E}_P[\hat{r}]=\mathbb{E}_{P'}[\hat{r}]$ {\em only if} $\sum_{i=1}^3p_ix_i=\sum_{i=1}^3p'_ix_i$. Hence, we need $\sum_{i=1}^3(p_i-p'_i)\mathbb{E}_{P_{x_i}}[x_i]=0$ {\em only if} $\sum_{i=1}^3(p_i-p'_i)x_i=0$. This implies that the null space of the matrix $[\mathbb{E}_{P_{x_1}}[x_1],\mathbb{E}_{P_{x_2}}[x_2],\mathbb{E}_{P_{x_3}}[x_3];1,1,1]$ is the same as the null space of the matrix $[x_1,x_2,x_3;1,1,1]$. Equivalently, the function that maps $x_i$ to $\mathbb{E}_{P_{x_i}}[x_i]$ is a linear function. By replacing $x_3$ with arbitrary $x$, it is easy to see that the function that maps $x_i$ to $\mathbb{E}_{P_{x_i}}[x_i]$ is the same for all chosen values of $\{x_i\}$. This completes the proof in this case.

For \textbf{Case 2}, if we consider a MAB instance with two arms with distributions $P,P'$ that witness the property in \textbf{Case 2}, then even if we have infinite samples from the quantization scheme we cannot achieve better than $O(|\mu_1-\mu_2|n)$ regret.

This shows that it suffices to consider quantization schemes that satisfy $(iii)$. We also note that: $(iv)$ for any $\delta>0$, the maximum distance between any consecutive quantization levels cannot exceed $1+\delta$, lest there is a reward $r$, that is in the middle of the two quantization levels, mapped to $\hat{r}$ with $|\hat{r}-r|\geq \frac{1+\delta}{2}$ which violates $(ii)$.

We are now ready to prove 1. Suppose towards a contradiction that $\exists b, t$ such that $\mathbb{P}[B_n\leq b]=1 \forall n>t$. Pick $n$ arbitrary large, we have that $b$ can describe at most $2^b$ quantization levels. As shown previously, we have that the maximum distance between quantization levels is bounded by a value $1+\delta$ for any $\delta>0$. Hence, either the interval $(-\infty,-2(2^b+1)]$ or the interval $[2(2^b+1),\infty)$ will have no quantization levels. We assume without loss of generality that the interval $[2(2^b+1),\infty)$ has no quantization levels. Hence, all the values in that interval will be mapped to values in $(-\infty,2(2^b+1))$. If the distribution of the reward is Gaussian for example, then the interval $(-\infty,2(2^b+1))$ will have non-zero probability. This contradicts $(iii)$.

To prove 2 we consider a gaussain reward distribution with zero mean and unit variance. We start by showing that to minimize the expected number of bits while satisfying $i,ii,iii$, the distance between the quantization levels need to be $1$ with one quantization level at $0$. 
Let $a\in \mathcal{Q}$ be the quantization level that is represented with the least number of bits, where $\mathcal{Q}$ is the set of quantization levels. We note that to minimize the average number of bits, we need to maximize the probability of transmitting a quantization level represented by a small number of bits. Pick $r>a$ such that $r-a<1$, and $Q^+=\{q\in \mathcal{Q}|q>r\}, Q^-=\{q\in \mathcal{Q}|q\leq r, q\neq a\}$. To satisfy $(iii)$ we need $r=\mathbb{E}[\hat{r}|r]=ap_a+\sum_{q\in\mathcal{Q}}qp_q$, where $p_q$ is the probability of quantizing $r$ to $q$. Hence,
\begin{equation}\label{eq:pa}
p_a=\frac{\sum_{q\in\mathcal{Q}^+}(q-r)p_q-\sum_{q\in\mathcal{Q}^-}(r-q)p_q}{r-a}.
\end{equation}
Thus, to maximize $p_a$, we maximize $q$ for $q\in \mathcal{Q}^+$ and choose $p_q=0$ for $q\in\mathcal{Q}^-$. From $(iv)$ this happens when the distance between the quantization levels in $\mathcal{Q}^+$ is $1$. Applying the same argument for $r<a$, we get that to maximize $p_a$, we need the distance between the quantization levels in $\mathcal{Q}^-$ to be $1$. Moreover, due to $(ii)$ the optimal choice for $a$ is $0$, lest from the fact that the density of a Gaussian distribution increases as we approach the mean, we can replace values that are mapped to $a$ with higher probabilities with values that are closer to $0$ (hence having higher densities), thus increasing $p_a$. Applying the same logic, to optimize the probability of $a'$, the quantization level represented by the second minimum number of bits, we require the distance between the quantization levels to be $1$ and $a'=1$. The same logic can  be applied for the quantization level represented by the $k$-th minimum number of bits showing that the optimal set of quantization levels is the set of integer numbers. 

Pick $r\in \mathbb{R}$, and let $p^{(r)}_z$ be the probability of quantizing $r$ to $z$. Let $p_z$ denote the probability of transmitting the quantization level at $z$. Then, $p_z = \int_{-\infty}^{\infty} p^{(x)}_z \frac{1}{\sqrt{2\pi}}e^{-{x^2}/{2}} dx$. By observing that for $x>\frac{z}{2}$, $e^{-{x^2}/{2}}$ is decays at least as $e^{-{z^2}/{8}}$, while for $x\leq \frac{z}{2}$ we have that $p^{(x)}_z$ decays at least as $e^{-{z^2}/{8}}$ due to $(ii)$, we get that $p_z$ is exponentially decaying in $z$. Hence, the optimal prefix-free encoding for levels is to assign $1$ bit for the quantized value $0$, $2$ bits for the value $1$, $3$ bits for the value $-1$ and so on \cite{cover1999elements}.
Note that in this proof/theorem, as well as in the upper bound, we consider prefix-free codes, yet very similar arguments provide arguments for non-prefix free codes as well.   

By solving $(iii)$ and $\sum_{z=-\infty}^\infty p^{(r)}_z=1$ together we get that $p^{(r)}_{\floor{r}}= \ceil{r}-r+ \sum_{i=2}^\infty (i-1)p^{(r)}_{\ceil{r}+i}- \sum_{i=2}^{\infty} (i-1)p^{(r)}_{\floor{r}-i}$. 
Due to $(ii)$ we can bound the quantization levels' probabilities as 

\begin{equation}
    P_i\geq \int_{i-1}^i (|x-i|-\sum_{i=1}^{\infty}ie^{-2(\ceil{x}+i)^2}) \frac{1}{\sqrt{2\pi}}e^{-{x^2}/{2}} dx+\int_{i}^{i+1} (|x-i|-\sum_{i=1}^{\infty}ie^{-2(\floor{x}-i)^2}) \frac{1}{\sqrt{2\pi}}e^{-{x^2}/{2}} dx.
\end{equation}
By computing this bound for $p_{-3},...p_3$, we get that average number of bits lower bounded by $2.5$ bits.


\end{proof}

\section{Proofs of Corollaries \ref{cor1}, \ref{cor2}}\label{app:cor}
The expected regret bounds follow directly from Theorem~\ref{main_thm}. To bound the average number of bits used for the avg-pt, we only need to bound the decay rate of $\frac{1}{n}\sum_{t=1}^{n-1}\frac{R_t}{\sigma t}$.

\underline{\textbf{Corollary~\ref{cor1}:}}\\
From Theorem~\ref{main_thm} and \cite{auer2002finite}, we have that for $\qname$ with UCB, there is a constant $C$ such that $R_n\leq C\sigma \sqrt{kn\log(n)}$. Then,
\begin{align}
    \frac{1}{n}\sum_{t=1}^{n-1}\frac{R_t}{\sigma t}&\leq C\frac{1}{n}\sum_{t=1}^{n}\frac{\sqrt{kt\log(t)}}{t}\nonumber \\
    &\leq \frac{C\sqrt{k\log(n)}}{n}\sum_{t=1}^{n}\frac{1}{\sqrt{t}}\nonumber \\
    &\leq \frac{C\sqrt{k\log(n)}}{n}(1+\int_{t=1}^n \frac{1}{\sqrt{t}})\nonumber \\
    &\leq C\sqrt{k\log(n)n}.
\end{align}

\underline{\textbf{Corollary~\ref{cor2}:}}\\
From Theorem~\ref{main_thm} and \cite{auer2002finite}, we have that for $\qname$ with $\epsilon$-greedy, there is a constant $C$ such that $R_n\leq C\sigma k\log(1+\frac{n}{k})$. Then, 
\begin{align}
    \frac{1}{n}\sum_{t=1}^{n-1}\frac{R_t}{\sigma t}&\leq \frac{Ck}{n}\sum_{t=1}^{n-1}\frac{\log(1+t)}{t}\nonumber \\
    &\leq \frac{Ck\log(1+n)}{n}\sum_{t=1}^{n-1}\frac{1}{t}\nonumber \\
    &\leq \frac{Ck\log(1+n)}{n}(1+\int_{1}^{n-1}\frac{1}{t})\nonumber \\
    &\leq \frac{Ck(\log(1+n))^2}{n}.
\end{align}

\section{Stochastic Linear Bandits Assumptions and Proof of Lemma~\ref{lem1}}\label{app:contex_bandits}
We assume the following
\begin{enumerate}
    \item $1\leq \beta_1\leq ... \leq \beta_n$ ($\beta_t$ is intuitively the radius of $\mathcal{C}_t$).
    \item $\max_{t\in\{1,...,n\}}\sup_{a\in \mathcal{A}_t}\langle\theta_t-\theta_*,a \rangle\leq \sqrt{\beta_n}$.\label{assump_2_app}
    \item $\|a\|_2\leq L\forall a\in \cup_{t=1}^n\mathcal{A}_t$.
    \item With probability at least $1-\frac{1}{n}$, for all $t\in \{1,...,n\}, \theta_*\in \mathcal{C}_t$.
\end{enumerate}
For rigorous description and justification of these assumptions please refer to \cite{lattimore2020bandit}. 

\underline{\textbf{Lemma~\ref{lem1}:}}\\
We observe that by Cauchy–Schwarz and assumption~\ref{assump_2_app}
\begin{align}\label{eq:int_stoch}
    \sum_{t=1}^{n}|\langle \theta_t-\theta_*, A_t \rangle|&\leq \sqrt{n \sum_{t=1}^{n}|\langle \theta_t-\theta_*, A_t \rangle|^2}\nonumber \\
    &\leq \sqrt{n\sum_{t=1}^{n}\min\{\beta_n,\langle \theta_t-\theta_*,A_t \rangle^2 \}}.
\end{align}
The proof of the expected regret and average number of bits bounds then follows as in \cite[Theorem~19.2]{lattimore2020bandit} using Theorem~\ref{main_thm}.



\section{A Case Where $1$ Bit is Sufficient}\label{app:mot}
\begin{figure*}[t!]
  \centering
\subfigure[$\lambda=1$.]{\includegraphics[width=0.49\linewidth]{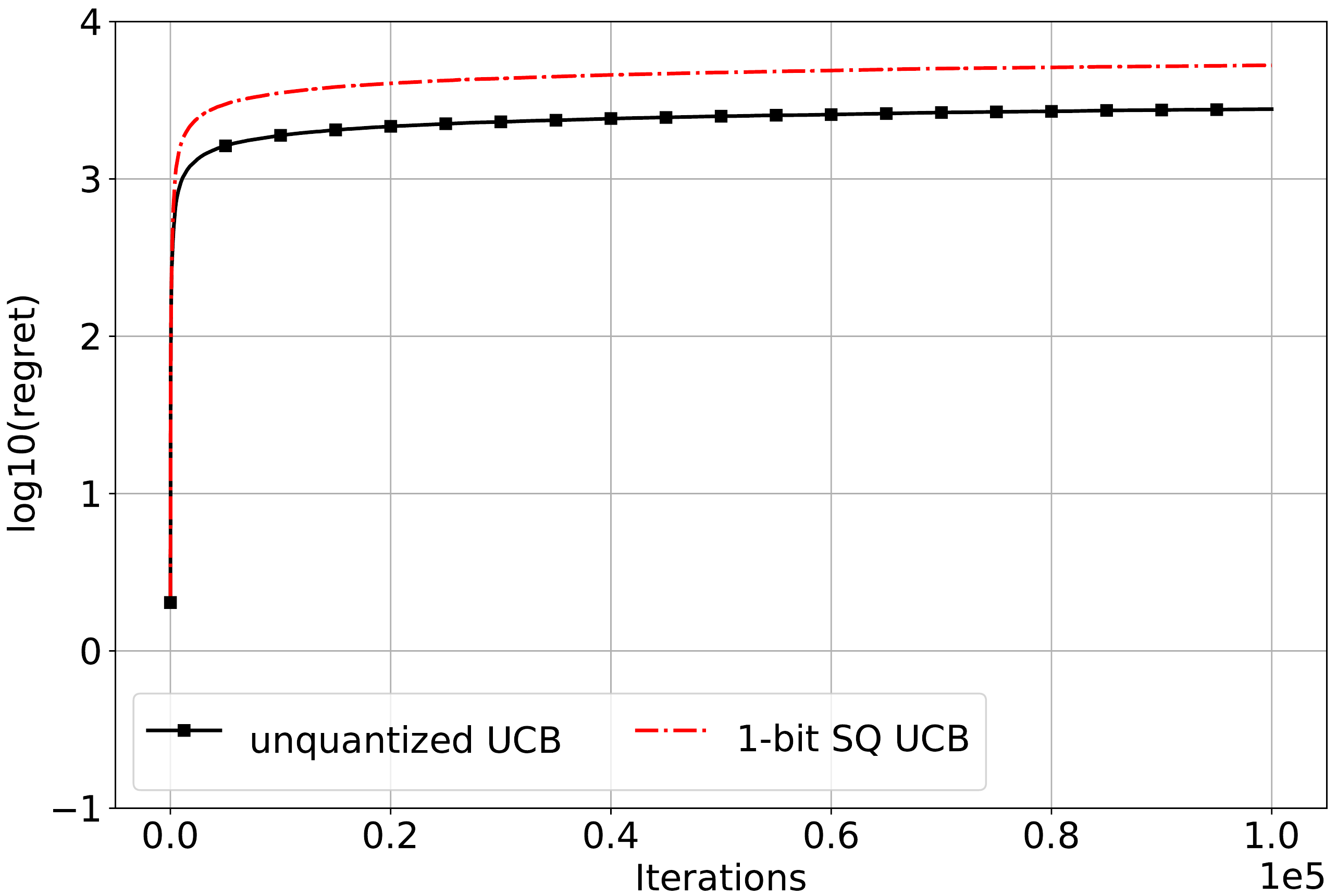}\label{fig:regrets_all_app_M1}}
 \subfigure[$\lambda =100$.]{\includegraphics[width=0.49\linewidth]{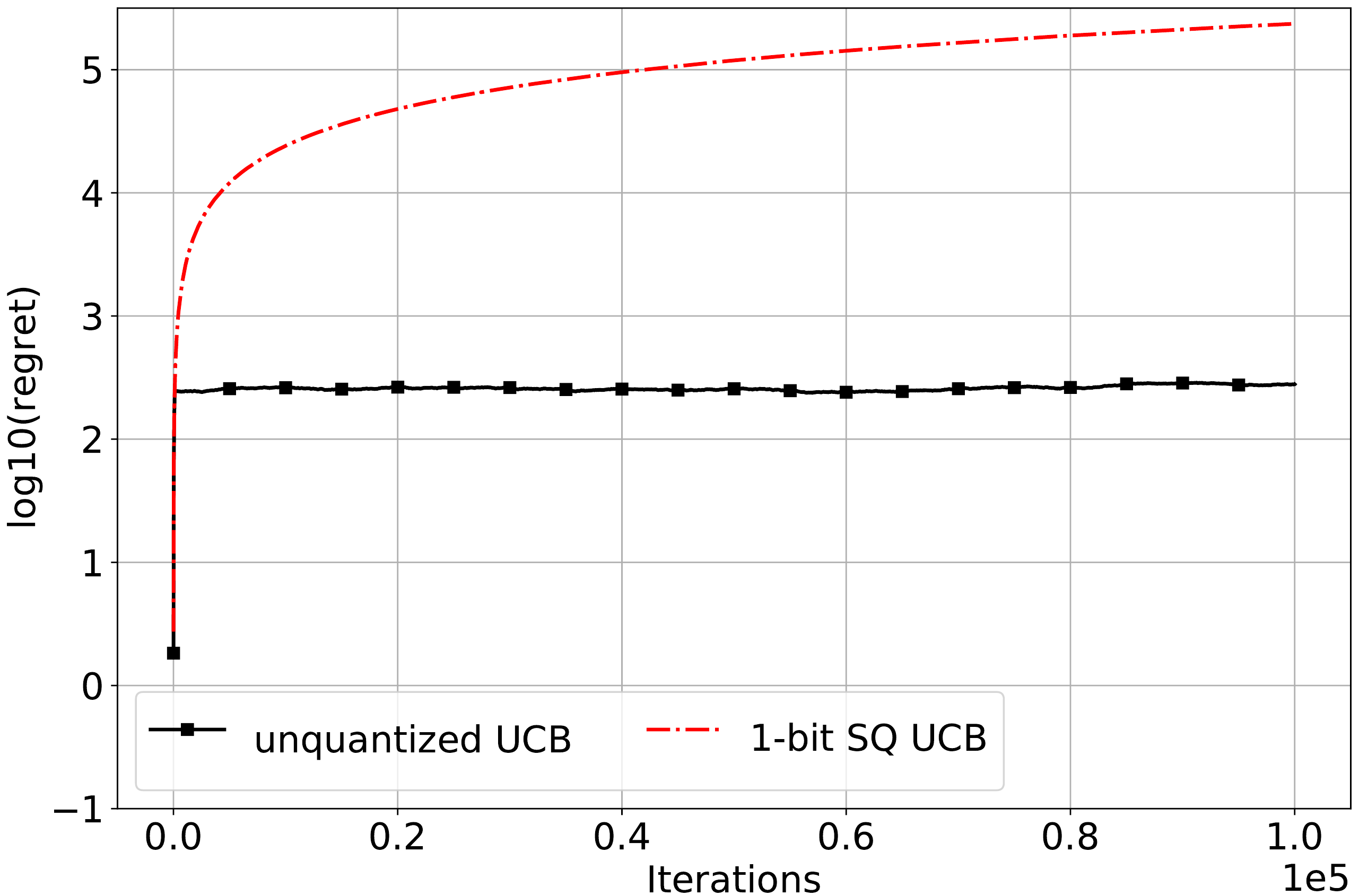}\label{fig:regret_UCB_app_M2}}
  \caption{Regret versus number of iterations.}\label{fig:iter_M_app}
\end{figure*}
In this appendix, we provide simulation results for the motivating example in Section~\ref{sec:example}. We consider the setup in Section~\ref{sec:eval} with $100$ arms where the arms' means are picked from a Gaussian distribution with mean $0$ and standard deviation $1$ and the reward distributions are conditionally Gaussian given the actions $A_t$ with variance $0.1$. The reward distributions are clipped to have support only on an interval $[-\lambda,\lambda]$. The parameter $\sigma_q$ is set to be $2$ when $\lambda=1$, and $0.1$ when $\lambda=100$ for the unquantized case. For $1$-bit SQ, $\sigma_q$ is set to be $2\lambda$. Fig.~\ref{fig:iter_M_app} shows the regret of unquantized and $1$-bit SQ with the UCB algorithm for $\lambda=1,100$. As discussed in Section~\ref{sec:example}, we observe a regret penalty when $\lambda \gg \sigma$.

\underline{\textbf{UCB regret bounds:}}\\
The UCB regrets bounds that we use in Section~\ref{sec:example} follow directly from the case of reward distributions that are supported on $[0,1]$ in \cite{auer2002finite}. To see the regret bound for rewards supported on $[-\lambda,\lambda]$, we observe that the expected regret can be written as $R_n=\sum_{t=1}^n\mathbb{E}(\mu^*_t-r_t)=\sum_{t=1}^n\mathbb{E}(\mu^*_t-\hat{r}_t)=2\lambda\sum_{t=1}^n\mathbb{E}(\frac{(\mu^*_t+\frac{1}{2})-(\hat{r}_t+\frac{1}{2})}{2\lambda})$, which transforms the problem to one with reward distributions supported on $[0,1]$.


\end{document}